\documentclass[onecolumn]{IEEEtranTIE}
\usepackage{graphicx}
\usepackage{cite}
\usepackage{picinpar}
\usepackage{amsmath}
\usepackage{amssymb}
\usepackage{url}
\usepackage{flushend}
\usepackage[latin1]{inputenc}
\usepackage{colortbl}
\usepackage{soul}
\usepackage{multirow}
\usepackage{caption}
\usepackage{pifont}
\usepackage{color}
\usepackage{alltt}
\usepackage[hidelinks]{hyperref}
\usepackage{enumerate}
\usepackage{siunitx}
\usepackage{breakurl}
\usepackage{epstopdf}
\usepackage{epsfig}
\usepackage{pbox}
\usepackage{float}
\usepackage{subfigure}
\usepackage{caption}
\usepackage[numbers,sort&compress]{natbib}
\usepackage{xr}  
\graphicspath{{images/}} 
\newtheorem{assumption}{Assumption}
\newtheorem{thm}{Theorem}
\newtheorem{rmk}{Remark}
\renewcommand{\IEEEQED}{\IEEEQEDclosed}
\begin{document}

\title{Model Predictive Control For Mobile Manipulators Based On Neural Dynamics\\
	(Extended version)}

\author{
	\vskip 1em
	
	Tao Su,
	Shiqi Zheng, \emph{Senior Member, IEEE}
}

\maketitle

\begin{abstract}
	This article focuses on the trajectory tracking problem of mobile manipulators (MMs). Firstly, we construct a position and orientation model predictive tracking control (POMPTC) scheme for mobile manipulators. The proposed POMPTC scheme can simultaneously minimize the tracking error, joint velocity, and joint acceleration. Moreover, it can achieve synchronous control for the position and orientation of the end-effector. Secondly, a finite-time convergent neural dynamics (FTCND) model is constructed to find the optimal solution of the POMPTC scheme. Then, based on the proposed POMPTC scheme, a non-singular fast terminal sliding model (NFTSM) control method is presented, which considers the disturbances caused by the base motion on the manipulator at the dynamic level. It can achieve finite-time tracking performance and improve the anti-disturbances ability. Finally, simulation and experiments show that the proposed control method has the advantages of strong robustness, fast convergence, and high control accuracy.
\end{abstract}


{}

\definecolor{limegreen}{rgb}{0.2, 0.8, 0.2}
\definecolor{forestgreen}{rgb}{0.13, 0.55, 0.13}
\definecolor{greenhtml}{rgb}{0.0, 0.5, 0.0}

\section{Introduction}

\IEEEPARstart{O}{ver} the past few decades, mobile manipulators (MMs) have played a powerful role in transportation, rescue, and military operations due to their mobility, agility, and flexibility\cite{article_1,article_2,article_28,article_29,article_32}. They can be used to perform a variety of tasks, such as rescue, assembling, and transportation. MMs are composed of a mobile base and manipulators, which combine the advantages of both and can work in very complex environments. Therefore, many researchers have conducted extensive research on MMs. Among them, trajectory tracking control is an important research direction in MMs, which can be divided into kinematic and dynamic control.

For kinematic control of MMs, the synchronous tracking of the position and orientation of the end-effector is very important. Recently, different trajectory tracking control methods \cite{article_3, article_4, article_5, article_6} of MMs have been proposed. For example, to solve the problem of joint angle drift of two redundant manipulators, \cite{article_7} proposed a new control scheme that can handle the repeat motion at the joint acceleration level. In order to ensure operational safety in dexterous tasks, joint angles, velocity, and acceleration need to be jointly constrained. However, most trajectory tracking methods for manipulators have limitations when dealing with multi-level joint constraints. For example, the minimum velocity norm (MVN) method \cite{article_8, article_9, article_10} and the minimum acceleration norm (MAN) method \cite{article_11} can handle various levels of joint constraints by applying transformation algorithms, but this may narrow the feasible region of control variables, thereby affecting the flexibility of the manipulator system. In addition, the motion of the base is also a factor that needs to be considered. \cite{article_12} used visual sensing to control a manipulator with a moving base. Simulation results showed that if the motion of the base was not considered, the end-effector could not fully track the trajectory. Therefore, how to achieve trajectory tracking while considering both joint constraints and the motion of the base is an important issue that needs to be addressed.

Model predictive control (MPC) is an advanced control method that can handle multiple constraints. However, due to the rolling optimization characteristics of MPC, it has some drawbacks, such as increased computational complexity and the existence of local optimal solutions. Therefore, it is crucial to solve the optimization problem efficiently and accurately. Lately, neural dynamics (ND) has proven to be effective in solving robot optimization problems \cite{article_13, article_14}. \cite{article_27} proposed a force-position control method for manipulators, which was based on the projection recurrent neural networks (PRNN).

For dynamic control of MMs, there are many excellent works, such as sliding model (SMC) and adaptive control. Among them, SMC has received widespread attention in the field of robot control due to its fast transient response and robustness to the disturbances \cite{article_15,article_16,article_17,article_18,article_30,article_33}. Traditional SMC uses a linear sliding surface, resulting in asymptotic tracking error convergence to zero in infinite time \cite{article_19}. To accelerate the convergence rate, \cite{article_20} proposed a terminal sliding mode (TSM) control method that utilized a nonlinear sliding surface. However, the TSM controller encounters singularity issues. To avoid singularity problems, \cite{article_21} constructed a non-singular terminal sliding mode (NTSM) using a power law approach. To further improve the convergence speed, \cite{article_19} and \cite{article_22} proposed a NFTSM control method, which could ensure rapid convergence near the equilibrium point. Due to its significant advantages, NFTSM methods have been used in many mechatronic systems, such as transmission systems, robotic manipulators, and electronic power systems \cite{article_23, article_24, article_25}.
However, the motion of the base can have a great influence on the manipulators during the operation process. In this case, the existing sliding mode control methods cannot achieve a satisfactory control performance. Therefore, how to achieve fast and stable control of manipulators under base motion is an another important problem to be solved.

In this article, a new MPC method based on neural dynamics is proposed for MMs. It can make the end-effector of the MM to stably track the desired trajectory when subjected to disturbances caused by base motion. The contributions are as follows:

\begin{enumerate}[1)]
	\item For kinematic control of MMs, a position and orientation model predictive tracking control (POMPTC) scheme is proposed. It can simultaneously track the position and orientation of the end-effector while considering the base motion.
	\item The POMPTC scheme is capable of effectively minimizing the tracking errors, velocity norms, and acceleration norms simultaneously. Compared with the exsiting methods \cite{article_9, article_10}, the POMPTC can directly handle multi-level joint constraints (joint angle, velocity and acceleration) without compromising the feasible domain.
	\item For the POMPTC scheme, a finite-time convergent neural dynamics (FTCND) model is proposed. This model utilizes the Li function as the activation function, which can solve the optimization problem in finite time.
	\item For dynamic control of MMs, a non-singular fast terminal sliding mode (NFTSM) control method is presented. This method can increase the tracking performance by compensating for the inertial disturbances caused by the base motion on the manipulator at the dynamic level. 
\end{enumerate}

The remainder of this article is organized as follows. In Section \ref{section:2}, the kinematic and dynamic models of the MMs are presented. In section \ref{section:3}, a POMPTC scheme is constructed and transformed into a convex optimization problem. Subsequently, we introduces the FTCND model and provides a corresponding finite-time convergence analysis. Section \ref{section:4} details the design process of the NFTSM controller and proves its stability. Section \ref{section:5} shows simulations and experiments results. Finally, Section \ref{section:6} summarizes the entire article.

\section{MODELING AND PROBLEM FORMULATION}\label{section:2}

This section introduces the kinematic and dynamic modeling of MMs.
\subsection{Kinematic Modeling}

\begin{figure}[!t]
	\begin{minipage}[t]{\linewidth}
		\centering
		\includegraphics[scale=0.06]{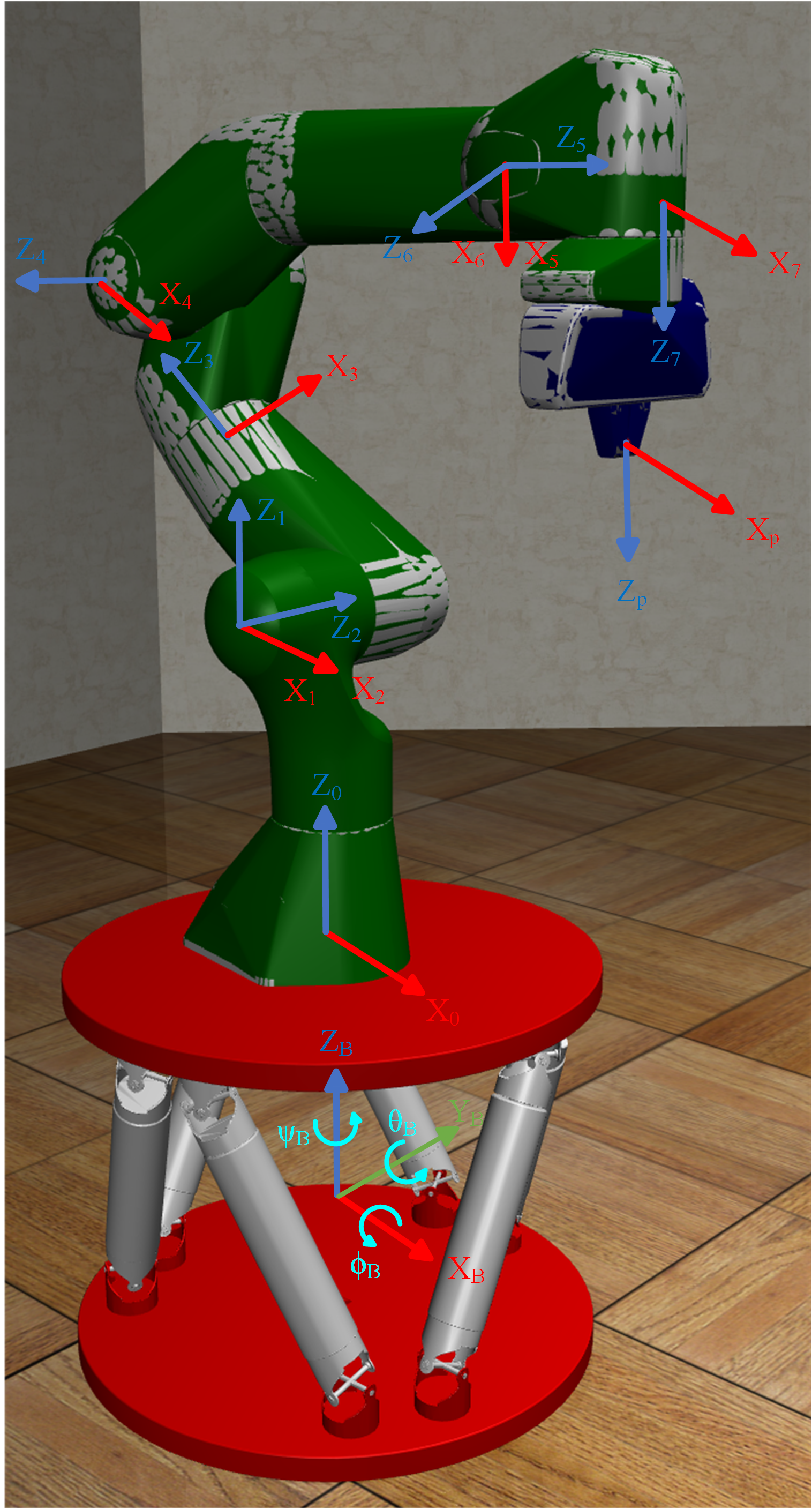}
	\end{minipage}
	\caption{Kinematics analysis of MMs}\label{FIG_1}
\end{figure}

In simple terms, MMs can be interpreted as a combination of a mobile platform and a manipulator. As shown in Fig.\ref{FIG_1}, to analyze the structure of MMs, mathematical modeling is conducted separately for the mobile platform and the redundant manipulator. Firstly, let $p = [\eta_{p_1}, \eta_{p_2}]^{\rm T}$ represent the coordinates of the end-effector, where $\eta_{p_1} = [x_p, y_p, z_p]$ and $\eta_{p_2} = [\phi_p, \theta_p, \psi_p]^{\rm T}$ denote the position and orientation relative to the inertial frame, expressed in Euler angles. The state variables of the system are defined as $q = [q_B, q_m]^{\rm T} \in \mathbb{R}^n$, where $q_B = [\eta_{B_1}, \eta_{B_2}]^{\rm T}$ includes the position $\eta_{B_1}$ and orientation $\eta_{B_2}$ of the base, and $q_m$ represents the vector of joint angles of the manipulator. More specifically, $\eta_{B_1} = [x_B, y_B, z_B]$ and $\eta_{B_2} = [\phi_B, \theta_B, \psi_B]^{\rm T}$ denote the position and orientation of the base relative to the inertial frame, also expressed in Euler angles. Therefore, there are the following representations:
\begin{equation}
	\label{eq1}
	\dot{q}_B=J_B(q_B)\rho,
\end{equation}
where $\rho$ represents the velocity of the base in the fixed coordinate system, and $J_B(q_B)$ represents the Jacobian matrix of the base. The generalized velocity of the end-effector is defined as $\dot{p} = [\dot{\eta}_{p_1}, \dot{\eta}_{p_2}]$, where $\dot{\eta}_{p_1}$ and $\dot{\eta}_{p_2}$ denote the linear and angular velocities of the end-effector, respectively. Additionally, the position and orientation of the end-effector relative to the inertial frame are determined by the forward kinematics of the entire system and are expressed as
\begin{equation}
	\label{eq2}
	p=F(q).
\end{equation} 

Furthermore, for the entire system, we have:
\begin{equation}
	\label{eq3}
	\dot{p}=\dot{F}(q)=J(q)\dot{q},
\end{equation}
where $\dot{q} = [\dot{q}_B, \dot{q}_m]^{\rm T} \in \mathbb{R}^n$ is the velocity vector, encompassing the velocity of the base relative to the inertial coordinate system as well as the joint velocities of the manipulator, and $J(q)$ is the geometric Jacobian matrix. It is noteworthy that $J(q)$ becomes singular at representation singularities, specifically when $\theta_B = \pm (\pi/2)$ and also when $det(J(q)[J(q)]^{\rm T}) = 0$. 

\subsection{Dynamic Modeling}

Assuming that the mass of the mobile base is significantly greater than the mass of the manipulator, and the motion of the base can be regarded as a disturbances to the manipulator. Therefore, the following system for an $n$-link rigid manipulator can be considered:
\begin{equation}
	\label{eq23}
	M(q_m)\ddot{q}_m+C(q_m,\dot{q}_m)\dot{q}_m +G(q_m)=\tau+\tau_d+\tau_b,
\end{equation}
where $\ddot{q}_m \in \mathbb{R}^n$ represents the joint acceleration of the manipulator. $\tau$ is the joint torque input, $\tau_d$ represents additional disturbances, and $\tau_b$ represents the disturbances caused by the base motion. $M(q_m) \in \mathbb{R}^{n\times n}$ is a positive-definite inertia matrix, $C(q_m,\dot{q}_m) \in \mathbb{R}^{n\times n}$ represents the centrifugal and Coriolis matrix, and $G(q_m) \in \mathbb{R}^n$ is the gravity vector of the rigid body model.
The tracking error is defined as $\widetilde{q}_1 = q_m - q_{md}$ and $\widetilde{q}_2 = \dot{q}_m - \dot{q}_{md}$, where $q_{md}$ is the desired joint angle and $\dot{q}_{md}$ is the desired joint angular velocity. From \eqref{eq23}, 
\begin{equation}
	\label{eq24}
	\begin{cases}
		\dot{\widetilde{q}_1}=\widetilde{q}_2, \\
		\dot{\widetilde{q}_2}=F(\widetilde{q})+M^{-1}(q_m)\tau+D(\widetilde{q},\dot{\widetilde{q}}) + B(\widetilde{q},\dot{\widetilde{q}}), \\
	\end{cases}.
\end{equation}
where $\widetilde{q} = \left[\widetilde{q}_1, \widetilde{q}_2\right]^{\rm T} \in \mathbb{R}^{2n}$, $F(\widetilde{q}) = -M^{-1}(q_m)(C(q_m,\dot{q}_m)\dot{q}\\+ G(q)) - \ddot{q}_{m_d}$, $D(\widetilde{q}, \dot{\widetilde{q}}) = -M^{-1}(q_m)\tau_d$, and $B(\widetilde{q},\dot{\widetilde{q}}) = -M^{-1}(q_m)\tau_b$.

\section{MOTION CONTROL DESIGN} \label{section:3}

This section constructs a POMPTC scheme to achieve trajectory tracking for MMs. Then, the POMPTC scheme is converted into an optimization problem, and a FTCND model is designed to solve this optimization problem. Theoretical analysis is conducted to prove that proves that the FTCND model has finite-time convergence in solving the QP problem.
\subsection{POMPTC Design}
The joint velocity $\dot{q}(j)$ is
\begin{equation}
	\label{eq5}
	\dot{q}(j)=\frac{q(j+1)-q(j)}{t},
\end{equation}
where $t$ represents the sampling period, and the joint acceleration $\ddot{q}(j)$ can be expressed as
\begin{equation}
	\label{eq6}
	\ddot{q}(j)\approx\frac{\dot{q}(j)-\dot{q}(j-1)}{t}=\frac{\Delta{\dot{q}(j)}}{t},
\end{equation}
thus
\begin{equation}
	\label{eq7}
	\begin{aligned}
		\dot{q}(j+i)&=\Delta{\dot{q}(j+i)}+\Delta\dot{q}(j+i-1)+\cdots+\\
		&\Delta\dot{q}(j)+\dot{q}(j-1),i=0,1,\cdots,N_u-1.
	\end{aligned}
\end{equation}
where $N_u$ represents the predictive control horizon. By combining equations \eqref{eq5} and \eqref{eq7}, the joint angle $q$ is expressed as
\begin{equation}
	\label{eq8}
	\begin{aligned}			
		q(j+i)&=q(j)+it\Delta{\dot{q}(j)}+(i-1)t\Delta\dot{q}(j+1)+\cdots+\\
		&t\Delta\dot{q}(j+i-1)+it\dot{q}(j-1),i=1,2,\cdots,N.
	\end{aligned}
\end{equation}

Subsequently, the POMPTC scheme can be constructed as
\begin{subequations}
	\label{eq10}
	\begin{align}
		\rm min \quad&\sum_{i=1}^{N}{||F(q(j+i))-p_d(j+i)||_{\mathcal{C}}^2}+ \nonumber\\ &\sum_{i=0}^{N_u-1}{||\dot{q}(j+i)||_{\mathcal{B}_1}^2}+\sum_{i=0}^{N_u-1}{||\Delta\dot{q}(j+i)||_{\mathcal{B}_2}^2}, \label{eq10:sub1}\\
		\rm s.t.\quad&{q_L\le{q(j+i)}\le{q_U},i=1,2,\cdots,N,} \label{eq10:sub2} \\
		\quad&{\dot{q}_L\le\dot{q}_(j+i)\le{\dot{q}_U},i=0,1,\cdots,N_u-1,} \label{eq10:sub3} \\
		\quad&{\ddot{q}_L\le\frac{\Delta\dot{q}_(j+i)}{t}\le{\ddot{q}_U},i=0,1,\cdots,N_u-1,} \label{eq10:sub4}
	\end{align}
\end{subequations}
where $||\cdot||_\mathcal{C} = \sqrt{(\cdot)^T\mathcal{C}}$ is the norm weighted by matrix $\mathcal{C}$, $p_d(j) \in \mathbb{R}^6$ represents the desired position and orientation. $\mathcal{C} \in \mathbb{R}^{6\times6}$ is the weight matrix for tracking errors, $\mathcal{B}_1 \in \mathbb{R}^{m\times m}$ is the weight matrix for joint velocities, $\mathcal{B}_2 \in \mathbb{R}^{m\times m}$ is the weight matrix for joint acceleration increments, and $\Delta\dot{q}(j) \in \mathbb{R}^m$ represents the increment in joint velocities. Here, $m$ denotes the number of joints in the robot. $q_L, \dot{q}_L, \ddot{q}_L$ are the minimum constraints for joint angles, velocities, and accelerations, respectively, while $q_U, \dot{q}_U, \ddot{q}_U$ are the corresponding maximum constraints. Subsequently, to simplify the above formulation, the following matrices are represented as
\begin{equation*}
	\begin{aligned}
		&U=\left[ 
		\begin{matrix}
			t & 0 & \cdots & 0 \\
			2t & t & \cdots & 0 \\
			\vdots & \vdots & \ddots & \vdots \\
			N_u\vdots & (N_u-1)t & \cdots & t \\
			\vdots & \vdots & \ddots & \vdots \\
			Nt & (N-1)t & \cdots & (N-N_u+1)t
		\end{matrix}
		\right] \in \mathbb{R}^{N\times{N_u}}, \\
		&J = J(q(j)) \in \mathbb{R}^{N_u\times{N_u}}, 
		I_1 = \left[ 
		\begin{matrix}
			1 & 0 & \cdots & 0 \\
			1 & 1 & \cdots & 0 \\
			\vdots & \vdots & \ddots & \vdots \\
			1 & 1 & \cdots & 1
		\end{matrix}
		\right] \in \mathbb{R}^{N_u\times{N_u}}, \\
	\end{aligned}
\end{equation*}
\begin{equation*}
	\begin{aligned}
		&O=[t\dot{q}(j-1),\cdots,N_ut\dot{q}(j-1),\cdots,Nt\dot{q}(j-1)]^{\rm T} \in \mathbb{R}^{N\times{m}}, \\
		&D_p = [p(j),p(j),\cdots,p(j)]^{\rm T} \in \mathbb{R}^{N\times{6}}, \\
		&D_v = [\dot{q}(j-1),\dot{q}(j-1),\cdots,\dot{q}(j-1)]^{\rm T} \in \mathbb{R}^{N_u\times{m}}, \\
		&D_q = [q(j),q(j),\cdots,q(j)]^{\rm T} \in \mathbb{R}^{N\times{n}}, \\
		&\Delta{V_k} = [\Delta{\dot{q}(j)},\cdots,\Delta{\dot{q}(j+N_u-1)}]^{\rm T} \in \mathbb{R}^{N_u\times{m}}, \\
		&V_k = [\dot{q}(j),\cdots,\dot{q}(j+N_u-1)]^{\rm T} \in \mathbb{R}^{N_u\times{m}}, \\
		&Q_k = [q(j+1),\cdots,q(j+N)]^{\rm T} \in \mathbb{R}^{N\times{m}}, \\
		&P_k = [p(j+1),\cdots,p(j+N)]^{\rm T} \in \mathbb{R}^{N\times{6}}, \\
		&P_d = [p_d(j+1),\cdots,p_d(j+N)]^{\rm T} \in \mathbb{R}^{N\times{6}}, \\
		&Q_U = [q_U,q_U,\cdots,q_U]^{\rm T} \in \mathbb{R}^{N\times{m}}, \\
		&Q_L = [q_L,q_L,\cdots,q_L]^{\rm T} \in \mathbb{R}^{N\times{m}}, \\
		&\dot{Q}_U = [\dot{q}_U,\dot{q}_U,\cdots,\dot{q}_U]^{\rm T} \in \mathbb{R}^{N_u\times{m}}, \\
		&\dot{Q}_L = [\dot{q}_L,\dot{q}_L,\cdots,\dot{q}_L]^{\rm T} \in \mathbb{R}^{N_u\times{m}}, \\
		&\ddot{Q}_U = [\ddot{q}_U,\ddot{q}_U,\cdots,\ddot{q}_U]^{\rm T} \in \mathbb{R}^{N_u\times{m}}, \\
		&\ddot{Q}_L = [\ddot{q}_L,\ddot{q}_L,\cdots,\ddot{q}_L]^{\rm T} \in \mathbb{R}^{N_u\times{m}},
	\end{aligned}		
\end{equation*}

Based on the aforementioned matrices, let
\begin{align}
	S = &2[(J\otimes{U}^{\rm T}C(J\otimes{U}))+ \nonumber\\
	&(I_4\otimes{I_1})^{\rm T}B_1(I_4\otimes{I_1})+B_2], \\
	G = &2(J\otimes{U})^{\rm T}C[\text{vec}(D_p)+\text{vec}(OJ^{\rm T})- \nonumber\\
	&\text{vec}(P_d)]+2(I_4\otimes{I_1}^{\rm T}B_1\text{vec}(D_v)),
\end{align}
where $C = I_2 \otimes \mathcal{C}$ with $I_2 \in \mathbb{R}^{N\times N}$, $B_1 = I_3 \otimes \mathcal{B}_1$, $B_2 = I_3 \otimes \mathcal{B}_2$ with $I_3 \in \mathbb{R}^{N_u\times N_u}$, and $I_4 \in \mathbb{R}^{m\times m}$.

We can simplify the cost function \eqref{eq10:sub1} by omitting constant terms unrelated to $\Delta{V_k}$ and rewrite it in a compact form
\begin{equation}
	\label{eq13}
	\underset{\Delta{V_k}}{\rm min}\quad\frac{1}{2}\text{vec}(\Delta{V_k})^{\rm T}S\text{vec}(\Delta{V_k})+G^{\rm T}\text{vec}(\Delta{V_k}).
\end{equation}

Besides, \eqref{eq10:sub2}-\eqref{eq10:sub4} can be reformulated as
\begin{equation}
	\label{eq14}
	H\text{vec}(\Delta{V_k})\le{w},
\end{equation}
where $H = I_4 \otimes H' \in \mathbb{R}^{m(2N+4N_u)\times{mN_u}}$ and $w =\text{vec}(w') \in \mathbb{R}^{m(2N+4N_u)}$ with
\begin{equation*}
	\begin{aligned}
		&H' = [V,-V,I_1,-I_1,I_4,-I_4]^{\rm T}, \\
		&w' = \left[ 
		\begin{matrix}
			Q_U-D_q-O \\
			-Q_L+D_q+O \\
			\dot{Q}_U-D_v \\
			-\dot{Q}_L+D_v \\
			t\ddot{Q}_U \\
			-t\ddot{Q}_L
		\end{matrix}
		\right],
	\end{aligned}
\end{equation*}

Then, convert \eqref{eq14} into a OP problem as follow:
\begin{subequations}
	\label{eq15}
	\begin{align}
		\underset{z(t)}{\rm min} \quad&\frac{1}{2}z(t)^{\rm T}Sz(t)+G^{\rm T}z(t), \\
		{\rm s.t.} \quad&Hz(t)\le{w}.
	\end{align}
\end{subequations}
of which $z(t) = \text{vec}(\Delta V_k) \in \mathbb{R}^{mN_u}$. At time $j$, the first element of the solution to the optimization problem \eqref{eq15}, namely $\Delta \dot{q}(j)$, can be used to calculate the joint velocity increment.

\subsection{FTCND Model Construction}

Design a slack variable vector $\varphi(t) \in \mathbb{R}^{m(2N+4N_u)}$ ($\varphi \geq 0$) to transform the OP problem \eqref{eq15} into a new form that incorporates the slack variables.
\begin{subequations}
	\label{eq16}
	\begin{align}
		\underset{z(t)}{\rm min} \quad&\frac{1}{2}z(t)^{\rm T}Sz(t)+G^{\rm T}z(t), \\
		{\rm s.t.} \quad&Hz(t)-w+\varphi(t)=0,
	\end{align}
\end{subequations}

Then, convert \eqref{eq16} into an unconstrained OP problem:
\begin{equation}
	\label{eq17}
	\begin{aligned}
		P(z(t),\varphi(t))=&\frac{1}{2}z(t)^{\rm T}Sz(t)+G^{\rm T}z(t)+ \\
		&\frac{1}{2}\xi||Hz(t)-w+\varphi(t)||^2,
	\end{aligned}
\end{equation}
where $\xi \in \mathbb{R}$ is a penalty factor. In addition, \eqref{eq17} is a convex OP problem, so there must be a solution that meets the conditions:
\begin{equation}
	\label{eq18}
	\begin{aligned}
		\begin{cases}
			&\nabla_{z}P=Sz^*+G+\xi{H^{\rm T}}(Hz^*-w+\varphi^*)=0, \\
			&\nabla_{\varphi}P=\xi(Hz^*-w+\varphi^*)=0,
		\end{cases}
	\end{aligned}
\end{equation}
where $z^*$ and $\varphi^*$ are the global optimal solutions. Therefore, \eqref{eq15} can be simplified to the following equation.
\begin{equation}
	\label{eq19}
	Nv(t)+D=0.
\end{equation}
with
\begin{equation*}
	\begin{aligned}
		N = \left[ 
		\begin{matrix}
			S+\xi{H^{\rm T}H} & \xi{H^{\rm T}} \\
			\xi{H} & \xi{I_4} \\
		\end{matrix}
		\right], 
		v(t)=\left[
		\begin{matrix}
			z(t) \\
			\varphi(t)
		\end{matrix}
		\right],
		D=\left[
		\begin{matrix}
			G-\xi{H^{\rm T}}w\\
			-\xi{w}
		\end{matrix}
		\right]
	\end{aligned}
\end{equation*}

After that, an error function is
\begin{equation*}
	h(t)=Nv(t)+D
\end{equation*}
when $h(t)$ tends to zero, the solution of \eqref{eq19} can be obtained. Therefore, this article adopts the ND formula with an activation function to make the error function approach zero, and it is designed as 
\begin{equation*}
	\dot{h}(t)=-\mu\Omega(h(t)),
\end{equation*}
where $\mu$ is an index related to the convergence rate, and $\Omega(\cdot)$ is an activation function composed of an odd and monotonically increasing $\omega(\cdot)$, and $\omega(\cdot)$ is expressed as
\begin{equation*}
	\omega(h_i)=\frac{1}{2}\lambda({\rm Lip}^\kappa(h_i)+{\rm Lip}^{1/\kappa}(h_i)))+\frac{1}{2}\zeta{h_i},
\end{equation*}
where $h_i$ represents the $i$-th element of $h(t)$, $\kappa\in(0,1)$ is a design parameter, $\lambda$ and $\zeta$ are positive parameters; and
\begin{equation*}
	{\rm Lip}^{\kappa}(h_i)=
	\begin{cases}
		|h_i|^{\kappa},	&h_i>0 \\
		0,	&h_i=0 \\
		-|h_i|^{\kappa}.	&h_i<0
	\end{cases},
\end{equation*}

Therefore, the solution of \eqref{eq19} is as follows:
\begin{equation}
	\label{eq20}
	N\dot{v}(t)=-\mu\Omega(Nv(t)+D).
\end{equation}

The finite time stability analysis is shown in Appendix~\ref{app:finite_time} in the supplementary file.

\section{NFTSM CONTROLLER DESIGN} \label{section:4}

In this section, an NFTSM controller with a dynamic model of the system will be designed.

The disturbances generated by the movement of the base can greatly influence the stability of the MMs system. For an $n$-link manipulator, it is necessary to compensate for the acceleration $a_{b_n} \in \mathbb{R}^3$ generated by the base movement. $\tau_b$ is defined as follows:
\begin{equation*}
	\tau_b = \left[\tau_{b_1}, \tau_{b_2}, \cdots, \tau_{b_n}\right]^{\rm T},
\end{equation*}
with
\begin{equation*}
	\tau_{b_n} = \sum_{i=1}^{n} m_i {a_{b_i}} \frac{\partial T_i}{\partial q_{m_i}} l_n.
\end{equation*}
where $m_i$ is the mass of the link, $T_i \in \mathbb{R}^{3 \times 3}$ is the transformation matrix, and $l_n \in \mathbb{R}^3$ is the position of the center of gravity of link $n$ relative to the front joint coordinate system.

To achieve a faster convergence speed and better tracking performance, the NFTSM surface is designed as follows:
\begin{equation}
	\label{eq25}
	s=\widetilde{q}_1+\alpha{sat(\Psi_1)}|\widetilde{q}_1|^{r_1}+\beta{sat(\Psi_2)}|\widetilde{q}_2|^{r_2},
\end{equation}
of which $\alpha>0$, $\beta>0$, $1<r_2<2$, $r_1>r_2$, $s=\left[s_1,...,s_n\right]^{\rm T} \in \mathbb{R}^n$, $\widetilde{q}_1=\left[\widetilde{q}_{11},...,\widetilde{q}_{1n}\right]^{\rm T}$, $\widetilde{q}_2=\left[\widetilde{q}_{21}, ..., \widetilde{q}_{2n}\right]^{\rm T}$, $\Psi_1=diag(\widetilde{q}_{11}, ..., \widetilde{q}_{1n})$, $\Psi_2=diag(\widetilde{q}_{21}, ..., \widetilde{q}_{2n})$.

%
%

\begin{assumption}\label{assump:1}
	The aggregated uncertainty $D(\widetilde{q},\dot{\widetilde{q}})$ is bounded.
\end{assumption}
\begin{thm}
	Under the designed NFTSM surface given by \eqref{eq25}, in order to ensure that the system error state $\widetilde{q}_1$ converges to zero, the following control law needs to be designed:
\end{thm}
\begin{equation}
	\label{eq27}
	\tau = -M(q)(u_{eq}+u_{sw}),
\end{equation}
with
\begin{equation}
	\label{eq28}
	\begin{aligned}
		u_{eq}&=\frac{|\widetilde{q}_2|^{2-r_2}sign(\widetilde{q}_2)}{\beta{r_2}}\cdot(1+\alpha{r_1}|\widetilde{q}_1|^{r_1-1})+F(\widetilde{q}), \\
		u_{sw}&=c_1|s|^{r_3}sign(s)+c_2s,
	\end{aligned}
\end{equation}
where $c_1, c_2$ and $r_3$ are designed positive constants, with $0 < r_3 < 1$. The NFTSM is capable of converging the system state to zero in finite time, thus exhibiting a faster convergence rate. Additionally, it avoids the singular problem, making it highly suitable for controlling MMs.

To mitigate chattering, a saturation function $sat(s)$ is employed in various controllers in lieu of the sign function $sign(s)$, and $sat(s)$ is
\begin{equation}
	\label{eq29}
	sat(s)=
	\begin{cases}
		1 &s>\delta, \\
		s/{\delta} &|s|\le{\delta}, \\
		-1 &s<-\delta,
	\end{cases}~~,
\end{equation}
where $\delta>0$ is called the boundary layer.

The stability analysis is shown in Appendix~B in the supplementary file.
\begin{rmk}
	The comparison with existing similar control methods are given in Table~\ref{table_2}. Firstly, from the perspective of joint constraints, the proposed method and \cite{article_26,article_7} all consider three-level joint constraints, while other approaches only consider two-level joint constraints. However, only the proposed method can achieve control of the end-effector in any orientation. Secondly, in terms of reducing the feasible region, the proposed method and \cite{article_26} do not narrow the feasible region, while other methods make the feasible region smaller.
\end{rmk}

\begin{table*}[h]
	\renewcommand{\arraystretch}{1.2}
	\caption{COMPARISONS OF VARIOUS TRAJECTORY TRACKING METHODS}
	\centering
	\label{table_2}
	\resizebox{\linewidth}{!}{
		\begin{tabular}{c c c c c}
			\hline\hline \\[-3mm]
			\multicolumn{1}{c}{} & \multicolumn{1}{c}{Optimization level} & \multicolumn{1}{c}{\pbox{20cm}{Joint constraints variables}} & \multicolumn{1}{c}{Orientation control level} & \multicolumn{1}{c}{Narrow feasible region} \\[1.6ex] \hline
			The proposed method & Acceleration & Joint angle, Joint velocity, and Joint acceleration & Any & No \\
			\cite{article_26} & Acceleration & Joint angle, velocity, and acceleration & N/A & No \\
			\cite{article_7} & Acceleration & Joint angle, velocity, and acceleration & N/A & Yes \\
			\cite{article_9}, \cite{article_10} & Velocity & Joint angle and velocity & Single & Yes \\
			\hline\hline
		\end{tabular}
	}
\end{table*}

\section{SIMULATIONS AND EXPERIMENTS} \label{section:5}
In this section, we conduct simulations and experiments to verify the proposed results.

\subsection{Simulations}
It should be noted that all simulations are conducted on Pycharm 2022 software based on the MuJoCo physics engine, which is installed on a computer with Windows 11, AMD Ryzen 5-6600H, and 16.0GB RAM.

\begin{figure}[!t]
	\subfigure
	{
		\begin{minipage}[t]{.25\linewidth}
			\includegraphics[scale=0.11]{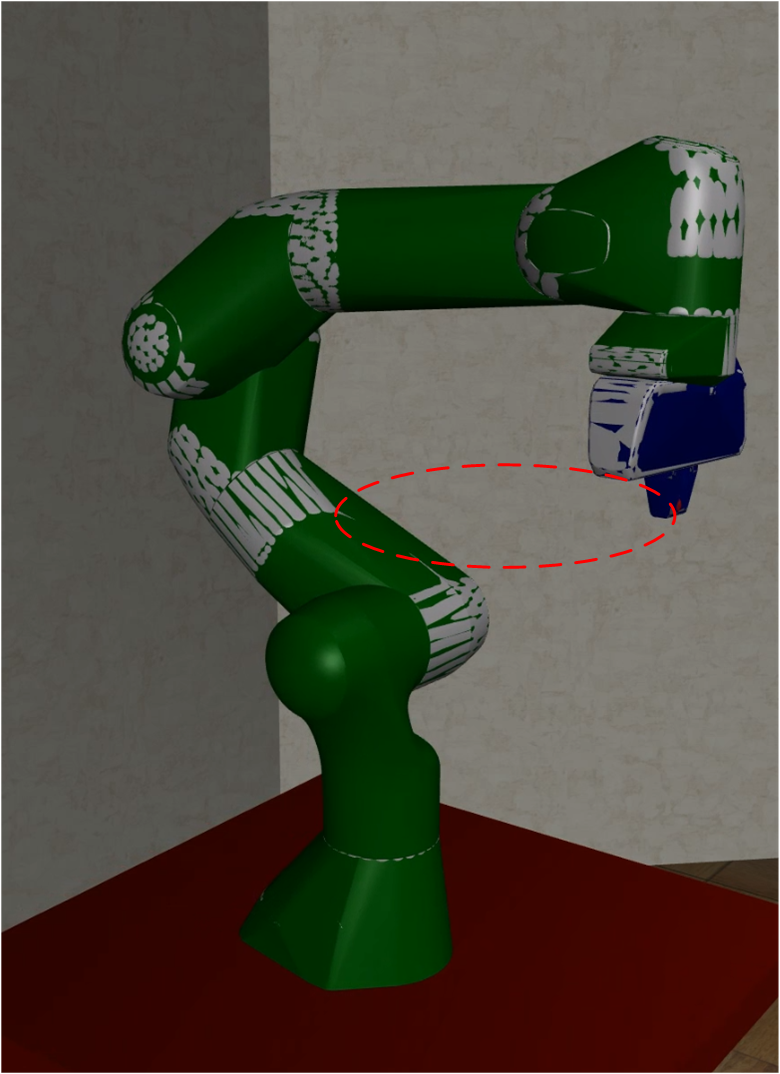}\\
			\centering{time=0s}
		\end{minipage}
		\begin{minipage}[t]{.25\linewidth}
			\includegraphics[scale=0.11]{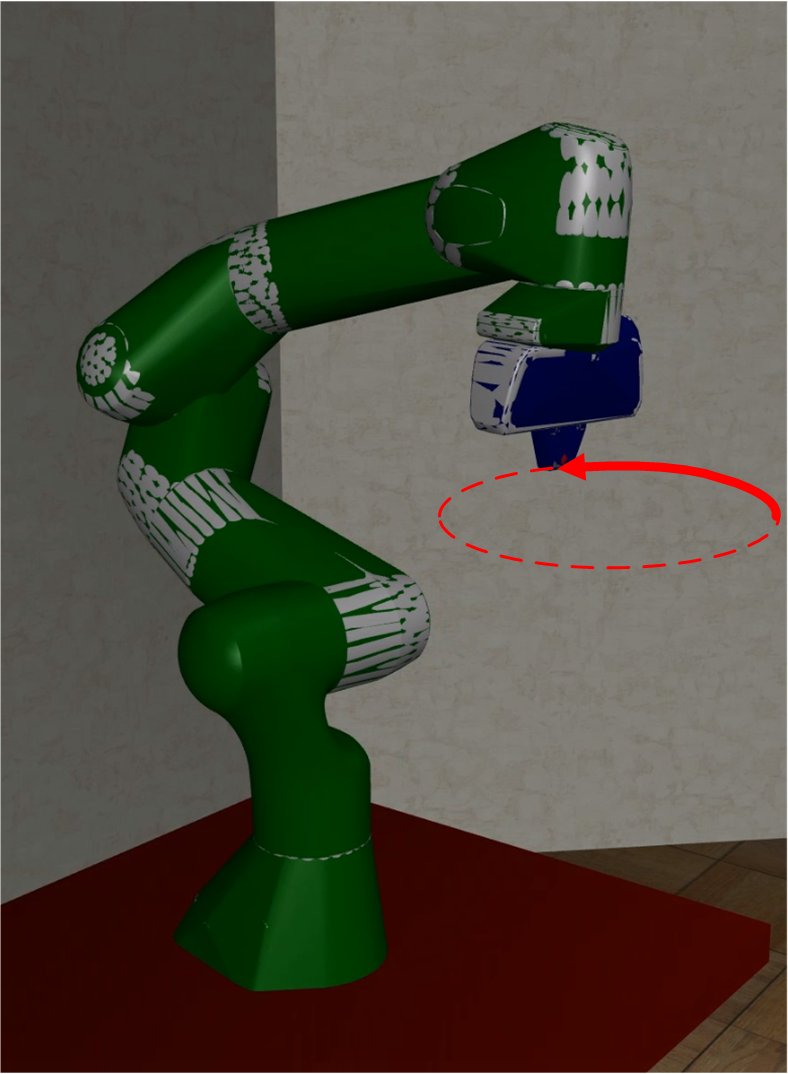}\\
			\centering{time=3s}
		\end{minipage}
		\begin{minipage}[t]{.25\linewidth}
			\includegraphics[scale=0.11]{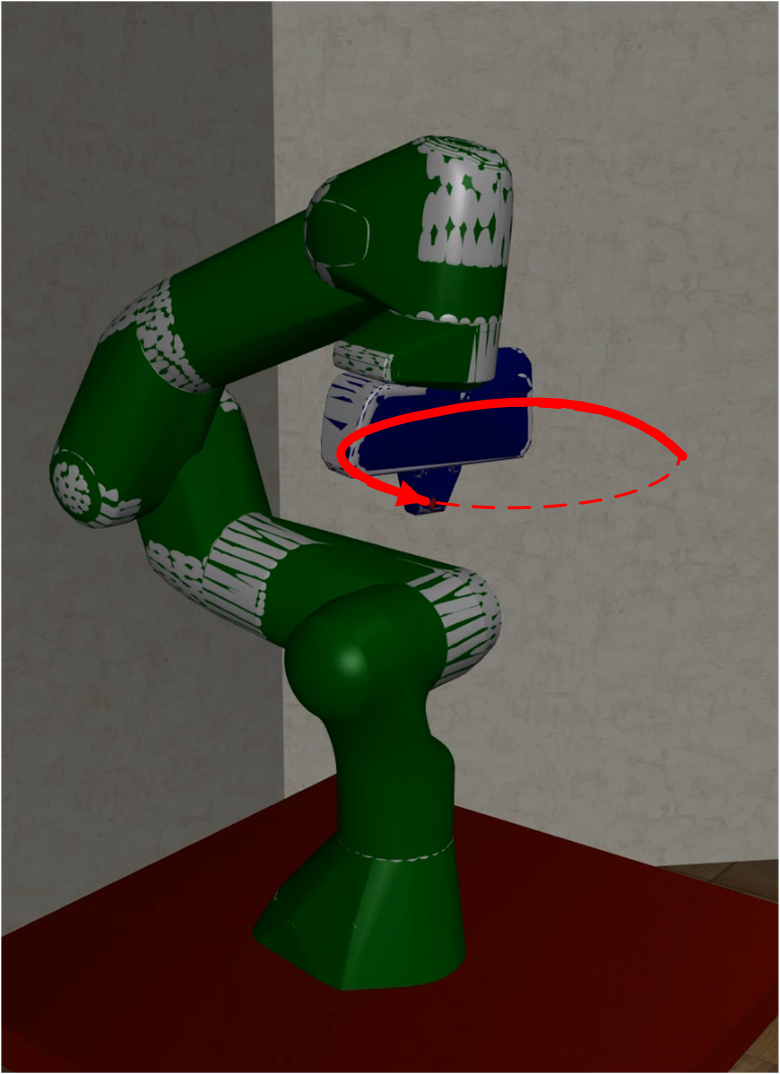}\\
			\centering{time=6s}
		\end{minipage}
		\begin{minipage}[t]{.25\linewidth}
			\includegraphics[scale=0.11]{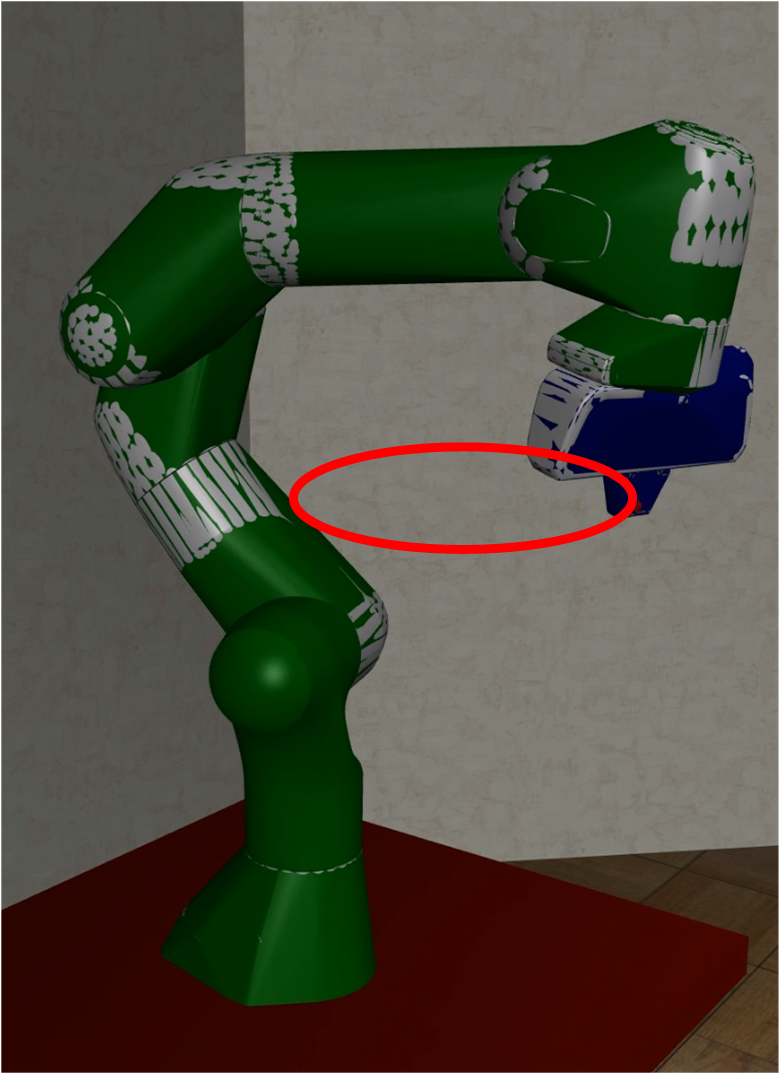}\\
			\centering{time=10s}
		\end{minipage}
	}
	\caption{MMs Simulation Model}\label{FIG_2}
\end{figure}

\begin{figure}[!t]
	\subfigure
	{
		\begin{minipage}[b]{.25\linewidth}
			\includegraphics[scale=0.26]{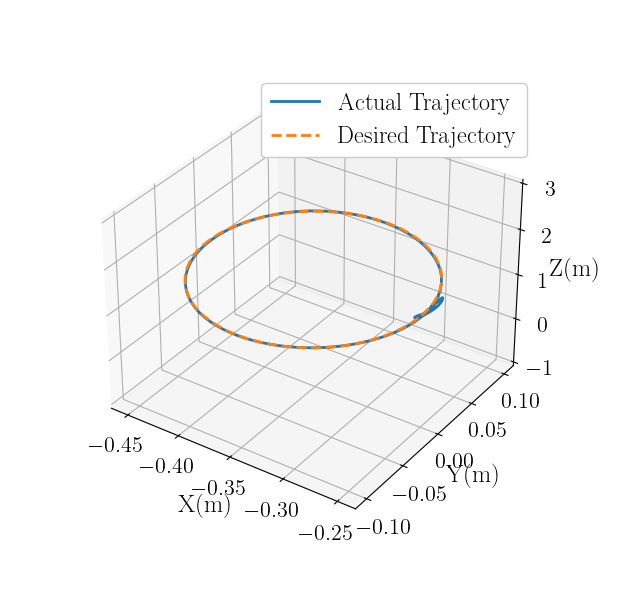}
			\centering{(a)}
		\end{minipage}
		\begin{minipage}[b]{.25\linewidth}
			\includegraphics[scale=0.23]{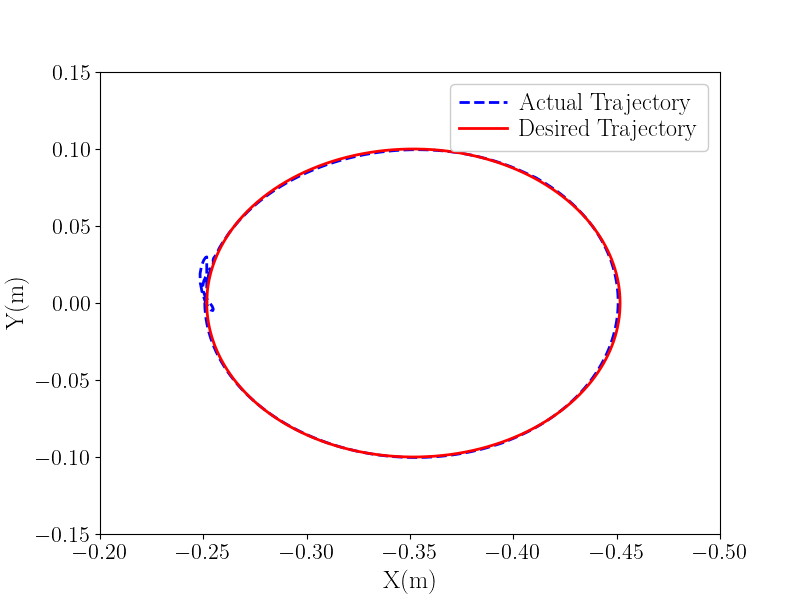} 
			\centering{(b)}
		\end{minipage}
		\begin{minipage}[b]{.25\linewidth}
			\includegraphics[scale=0.23]{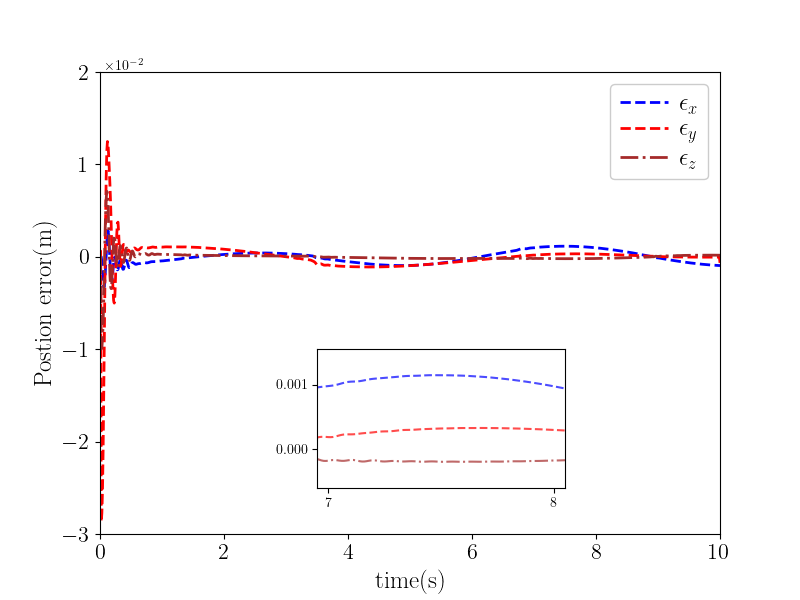}
			\centering{(c)}
		\end{minipage}
		\begin{minipage}[b]{.25\linewidth}
			\includegraphics[scale=0.23]{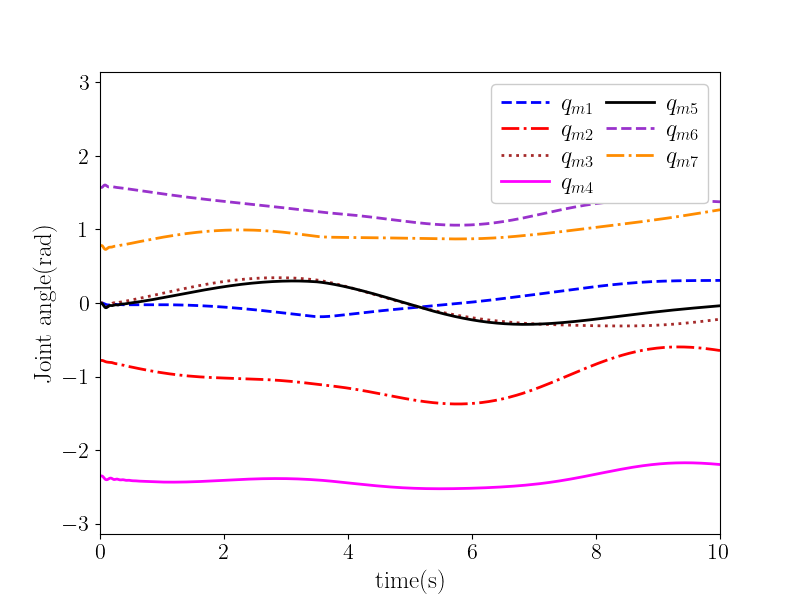}
			\centering{(d)}
		\end{minipage}				
	}
	\subfigure
	{
		\begin{minipage}[b]{.25\linewidth}
			\centering
			\includegraphics[scale=0.23]{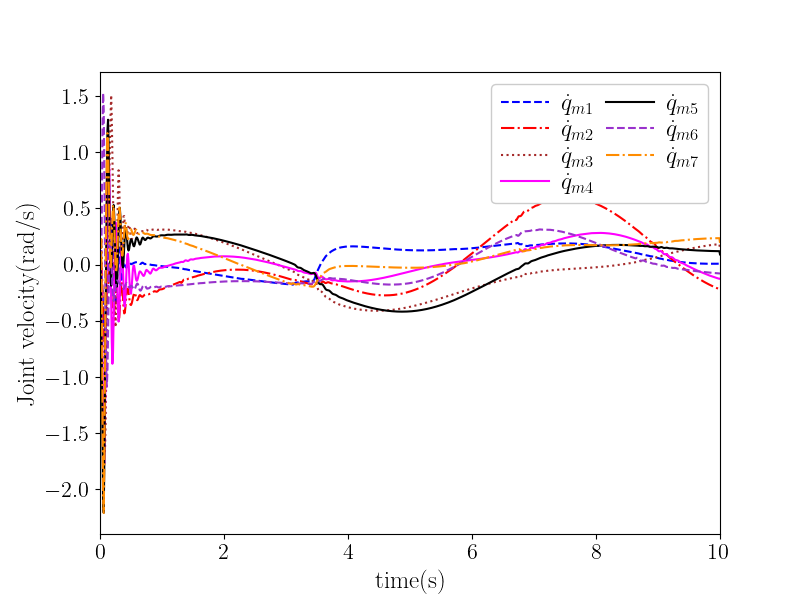} 
			\centering{(e)}
		\end{minipage}
		\begin{minipage}[b]{.25\linewidth}
			\centering
			\includegraphics[scale=0.23]{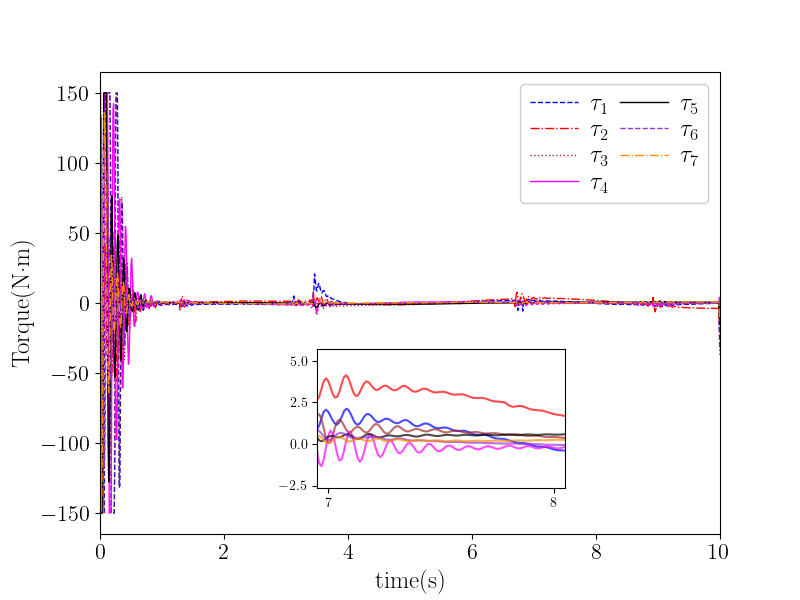}
			\centering{(f)}
		\end{minipage}
		\begin{minipage}[b]{.25\linewidth}
			\centering
			\includegraphics[scale=0.23]{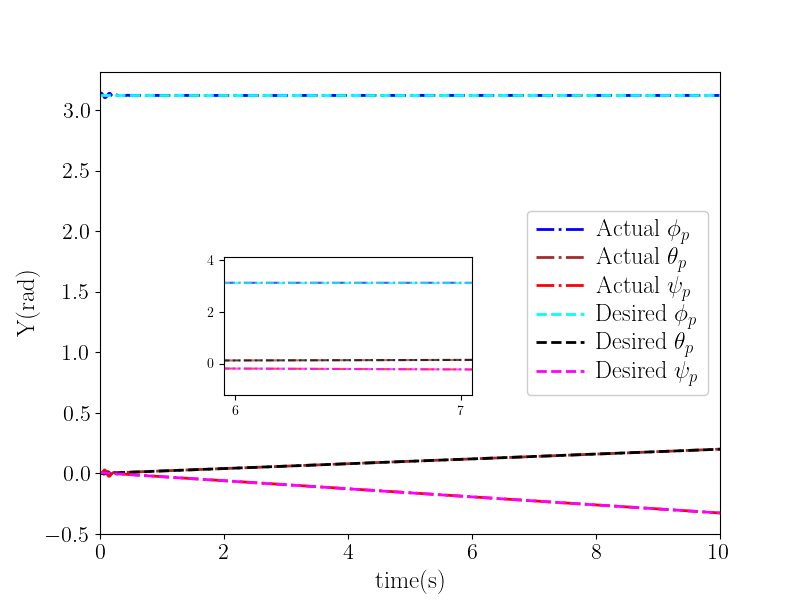}
			\centering{(g)}
		\end{minipage}
		\begin{minipage}[b]{.25\linewidth}
			\centering
			\includegraphics[scale=0.23]{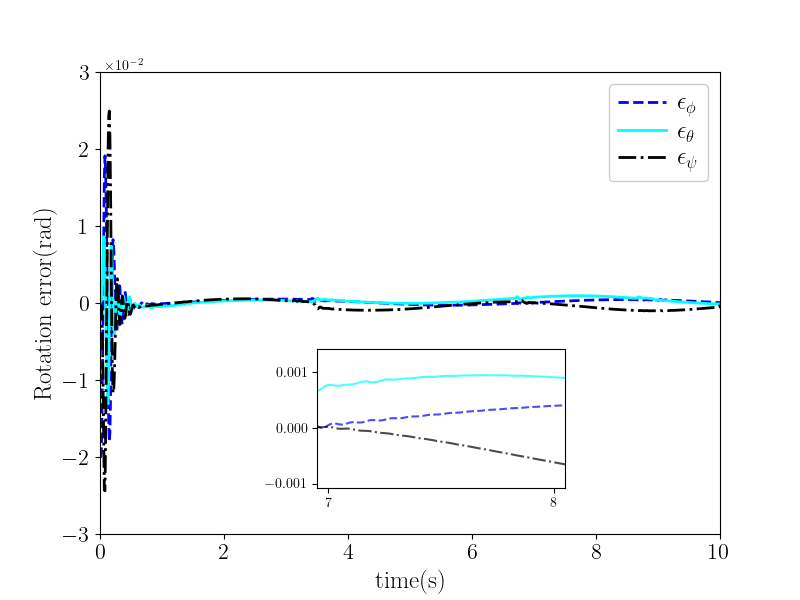}
			\centering{(h)}
		\end{minipage}
	}
	\caption{The simulation results show that the NFTSM algorithm combined with POMPTC strategy tracks the circular trajectory. (a) Three-dimensional desired trajectory and actual trajectory. (b) Two-dimensional actual trajectory and desired trajectory. (c) Position error of the end-effector. (d) Joint angle of the manipulator. (e) Joint velocity of the manipulator. (f) Joint torque. (g) Actual and desired rotation vector. (h) The errors of the rotation vector.}\label{FIG_4}
\end{figure}

In the simulation, the Franka Emika Panda, a 7-DOF redundant manipulator, is mounted on a Stewart platform. As is shown in Fig.~\ref{FIG_2}, the trajectory tracking task is to follow a circular path. The relevant parameters of the redundant manipulator are set to $m=13$, with the number of joints being $n=7$. The joint constraints are shown in Appendix~C in the supplementary file.

In addition, the base in the simulation is set as a six-degree-of-freedom massless square block, with the first three degrees of freedom controlling the direction of movement and the last three controlling the direction of rotation. The entire base is controlled independently.

In the simulation, the parameters of the POMPTC strategy are selected as $T=10$ s, $C=50000I_1$, $B_1=I_2$, $B_2=20I_2$, $N=N_u=5$, $\xi=5$, $\lambda=1$, $\zeta=30$, $\kappa=0.8$, $\mu=5$, and initial joint angle $q_0=[0, -0.78, 0, -2.35, 0, 1.57, 0.78]^{\rm T}~{\rm rad}$. The NFTSM parameters are selected as $\alpha=1$, $\beta=1$, $r_1=1.8$, $r_2=1.6$, $r_3=1$, $c_1=20$, $c_2=0.6$.

The simulation results of the NFTSM controller combined with MPC are shown in Fig.~\ref{FIG_4}. It can be seen from the figure that the position and direction tracking errors converge rapidly to 0.01 m and 0.01 rad respectively, and all joint angles and velocities remain within the limit range. Therefore, the proposed method is effective and has high control accuracy.

\begin{figure}[!t]
	\centering
	\subfigure
	{
		\begin{minipage}[t]{.33\linewidth}
			\includegraphics[scale=0.2]{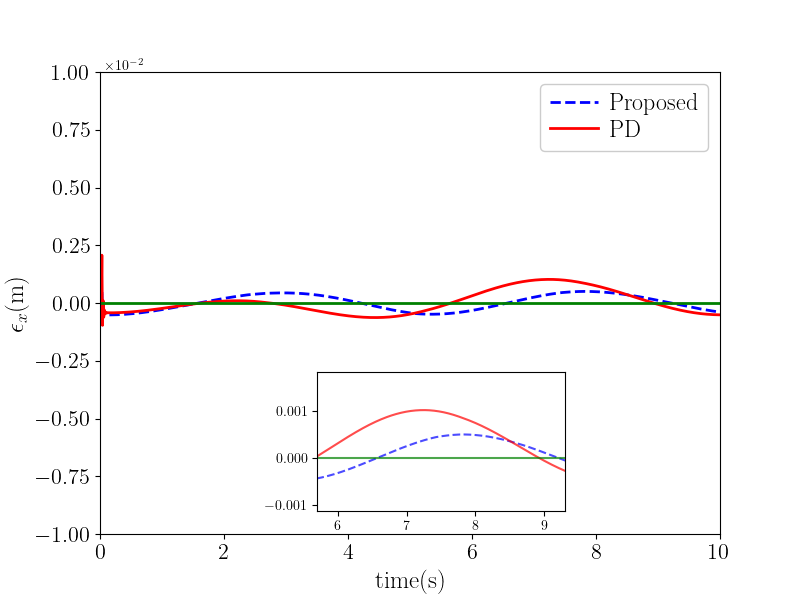}
		\end{minipage}
		\begin{minipage}[t]{.33\linewidth}
			\includegraphics[scale=0.2]{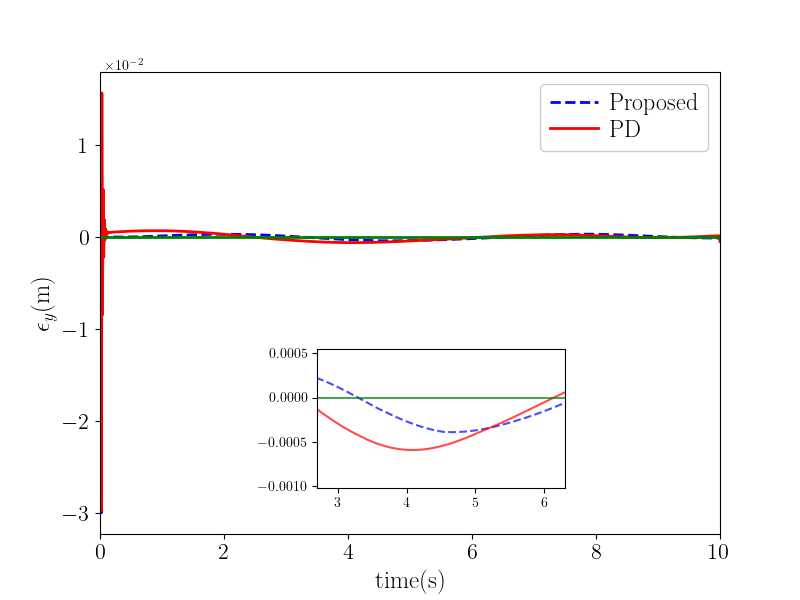}
		\end{minipage}		
		\begin{minipage}[t]{.33\linewidth}
			\includegraphics[scale=0.2]{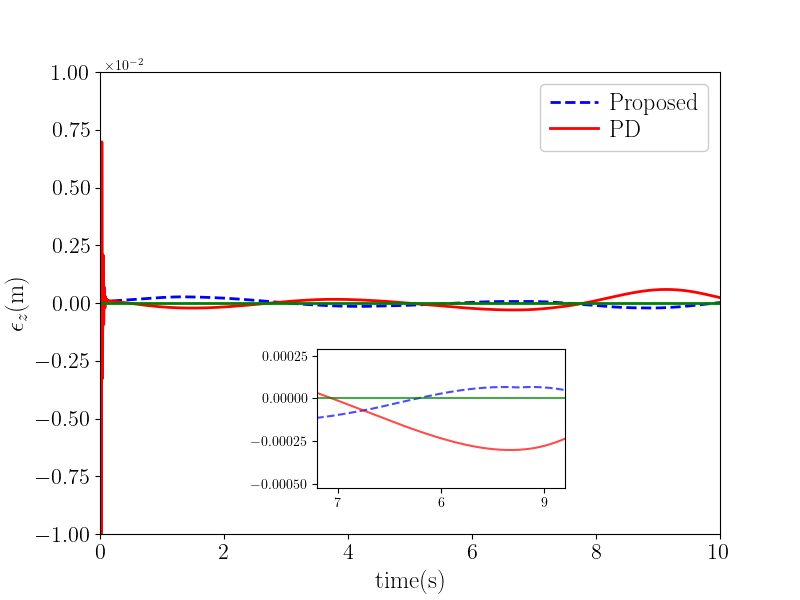}
		\end{minipage}
	}
	\subfigure
	{
		\begin{minipage}[t]{.33\linewidth}
			\includegraphics[scale=0.2]{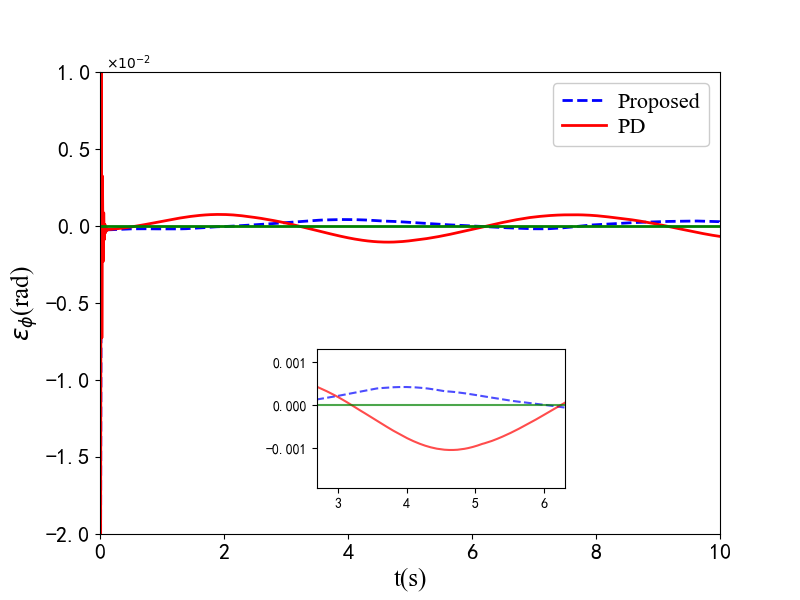}
		\end{minipage}
		\begin{minipage}[t]{.33\linewidth}
			\includegraphics[scale=0.2]{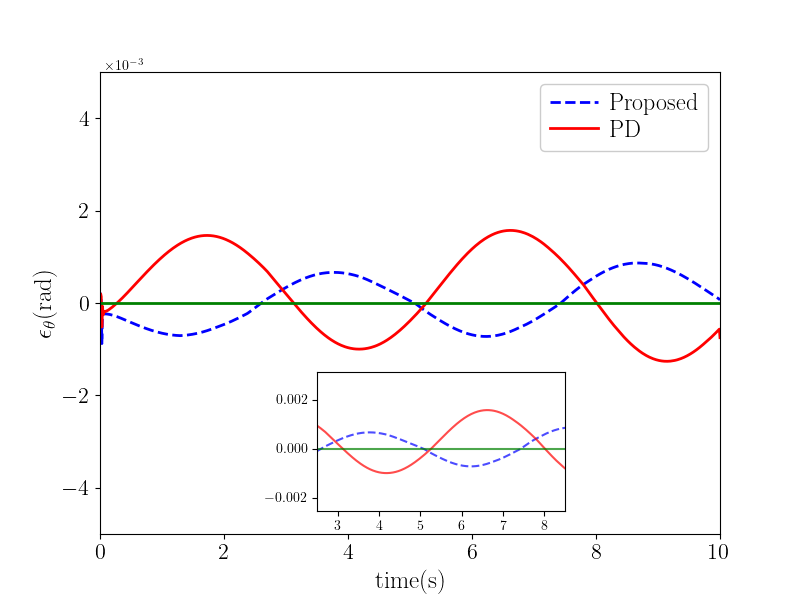}
		\end{minipage}		
		\begin{minipage}[t]{.33\linewidth}
			\includegraphics[scale=0.2]{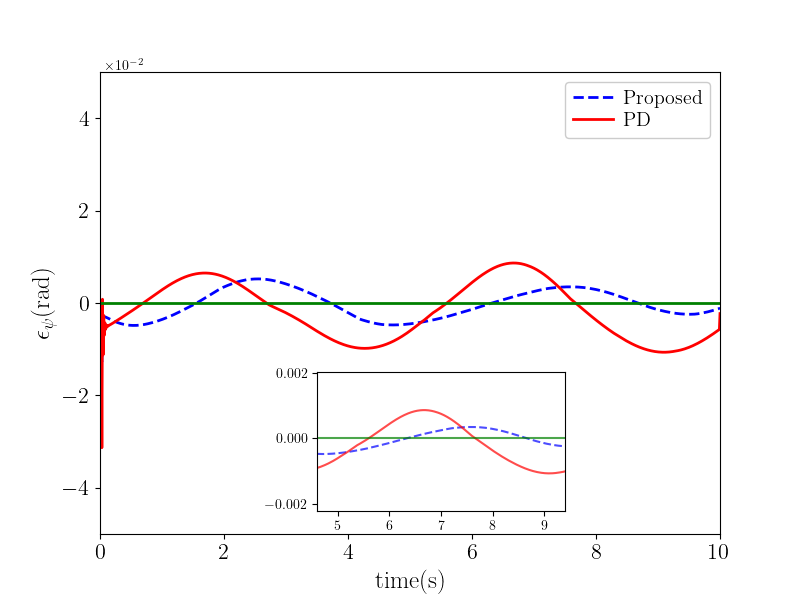}
		\end{minipage}
	}
	\caption{End position and rotation error of redundant manipulator under different motion control algorithms.}\label{FIG_5}
\end{figure}
\begin{figure}[!t]
	\centering
	\subfigure
	{
		\begin{minipage}[t]{.33\linewidth}
			\includegraphics[scale=0.2]{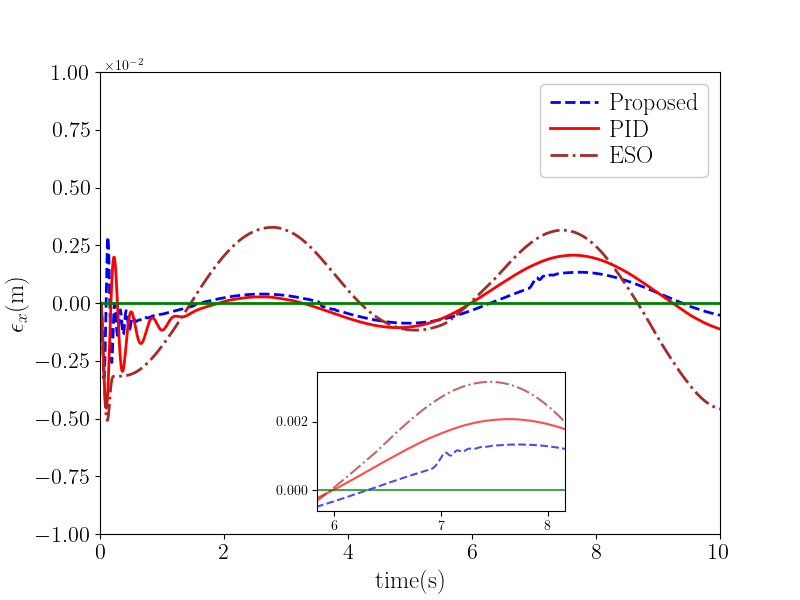}\\
			\centering{(a)}
		\end{minipage}
		\begin{minipage}[t]{.33\linewidth}
			\includegraphics[scale=0.2]{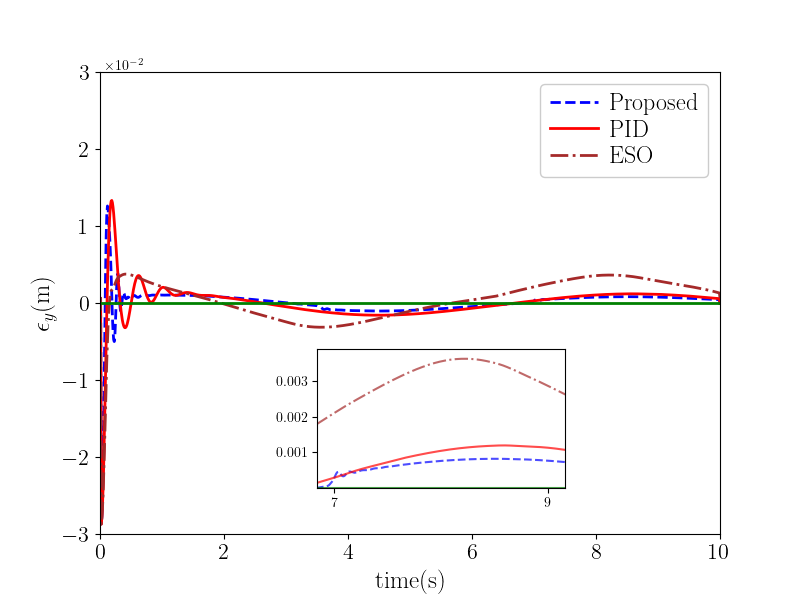}\\
			\centering{(b)}
		\end{minipage}		
		\begin{minipage}[t]{.33\linewidth}
			\includegraphics[scale=0.2]{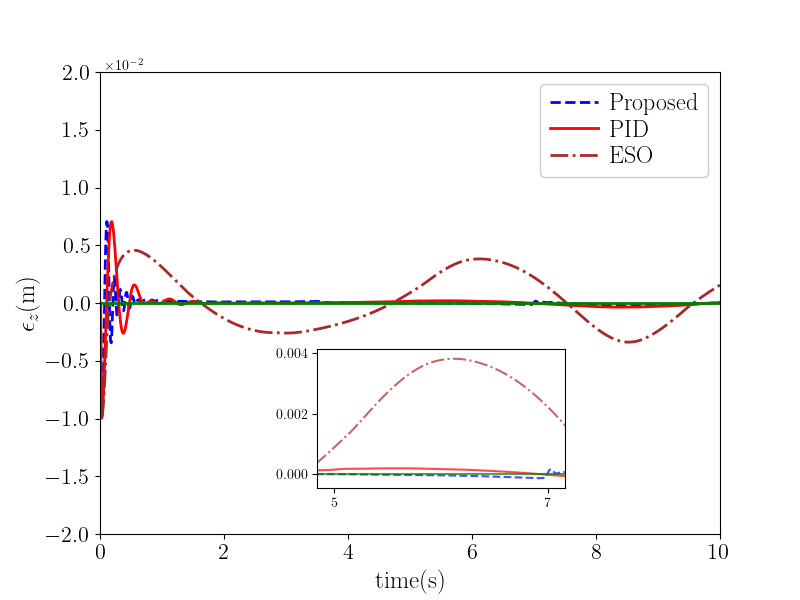}\\
			\centering{(c)}
		\end{minipage}
	}
	\caption{End position error of redundant manipulator under different control algorithms}\label{FIG_6}
\end{figure}

\begin{figure}[ht]
	\centering
	\subfigure
	{
		\begin{minipage}[t]{.25\linewidth}
			\includegraphics[scale=0.11]{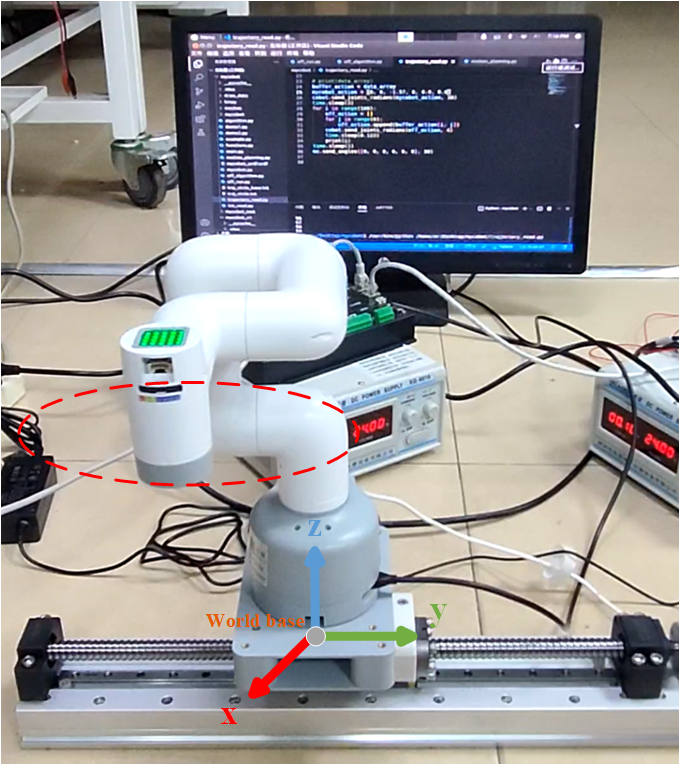}\\
			\centering{time=0s}
		\end{minipage}
		\begin{minipage}[t]{.25\linewidth}
			\includegraphics[scale=0.11]{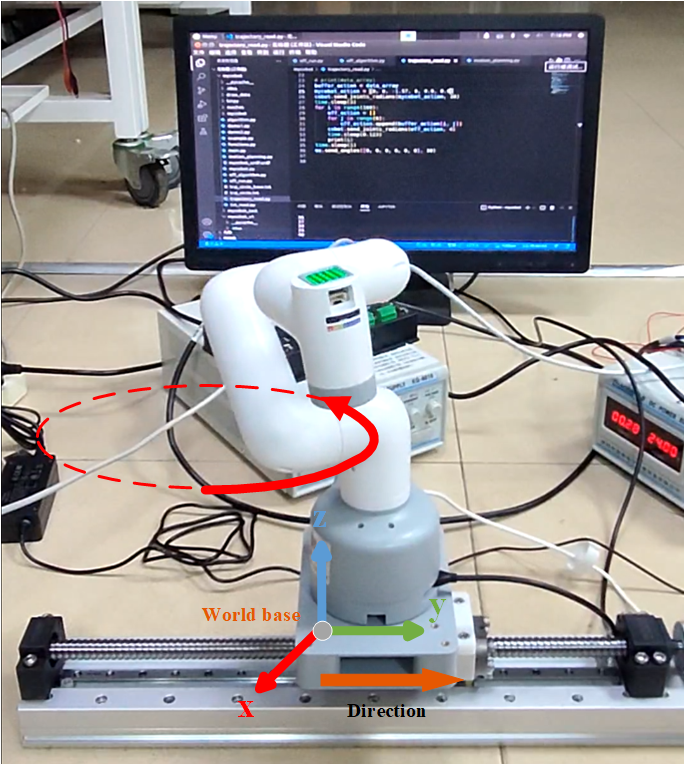}\\
			\centering{time=3s}
		\end{minipage}		
		\begin{minipage}[t]{.25\linewidth}
			\includegraphics[scale=0.11]{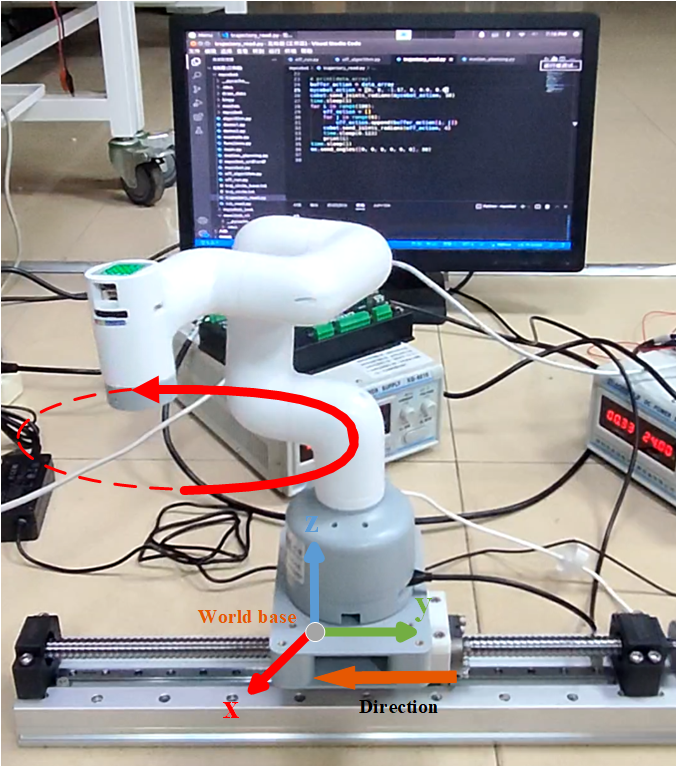}\\
			\centering{time=6s}
		\end{minipage}
		\begin{minipage}[t]{.25\linewidth}
			\includegraphics[scale=0.11]{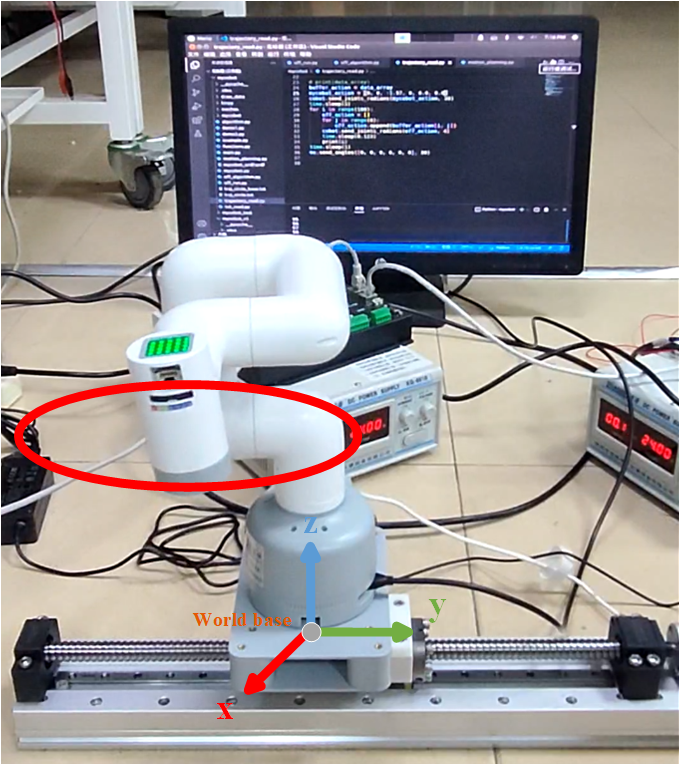}\\
			\centering{time=10s}
		\end{minipage}
	}
	\centering{(a)}
	\subfigure
	{
		\begin{minipage}[t]{.25\linewidth}
			\includegraphics[scale=0.11]{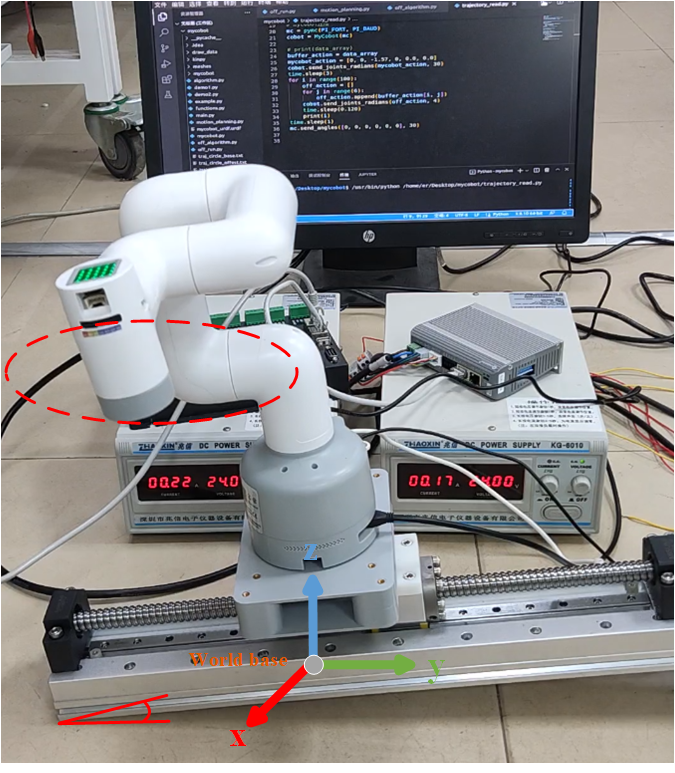}\\
			\centering{time=0s}
		\end{minipage}
		\begin{minipage}[t]{.25\linewidth}
			\includegraphics[scale=0.11]{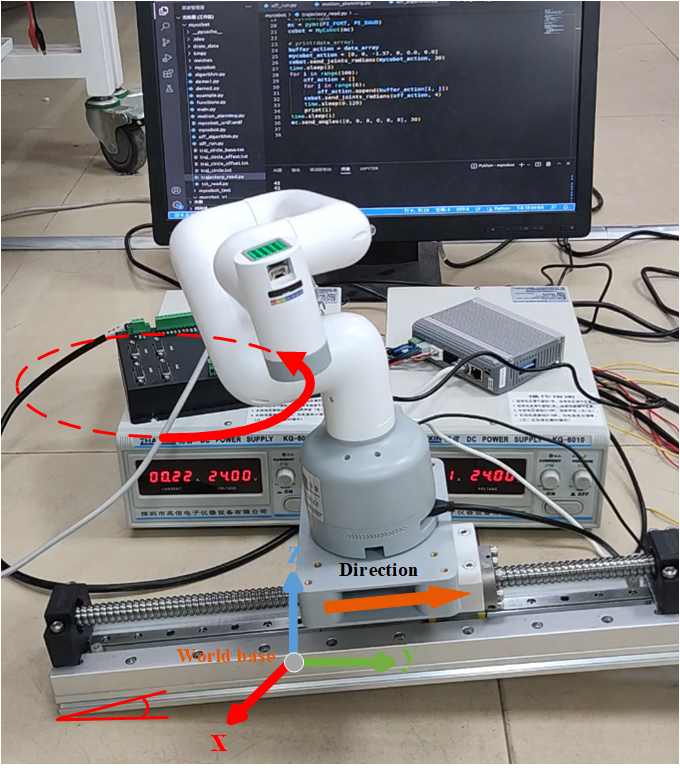}\\
			\centering{time=3s}
		\end{minipage}		
		\begin{minipage}[t]{.25\linewidth}
			\includegraphics[scale=0.11]{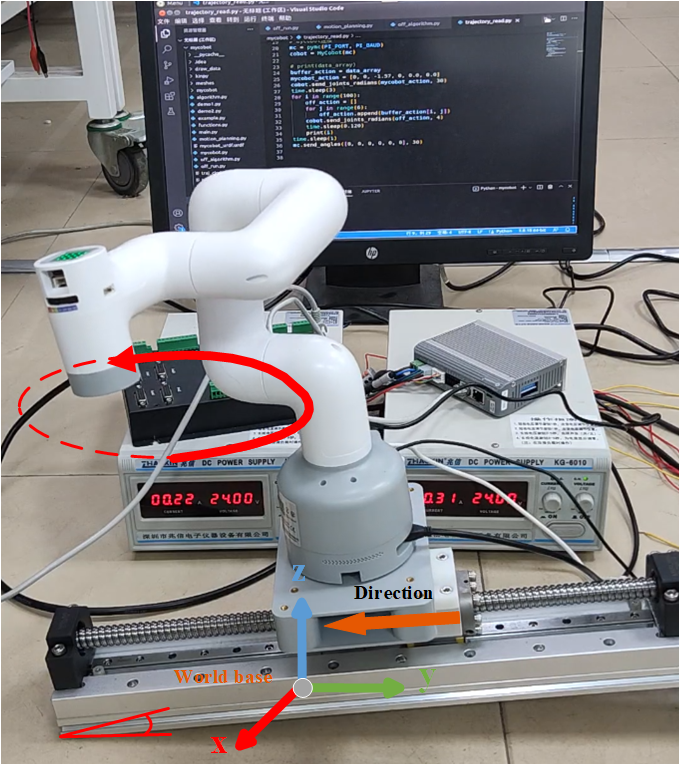}\\
			\centering{time=6s}
		\end{minipage}
		\begin{minipage}[t]{.25\linewidth}
			\includegraphics[scale=0.11]{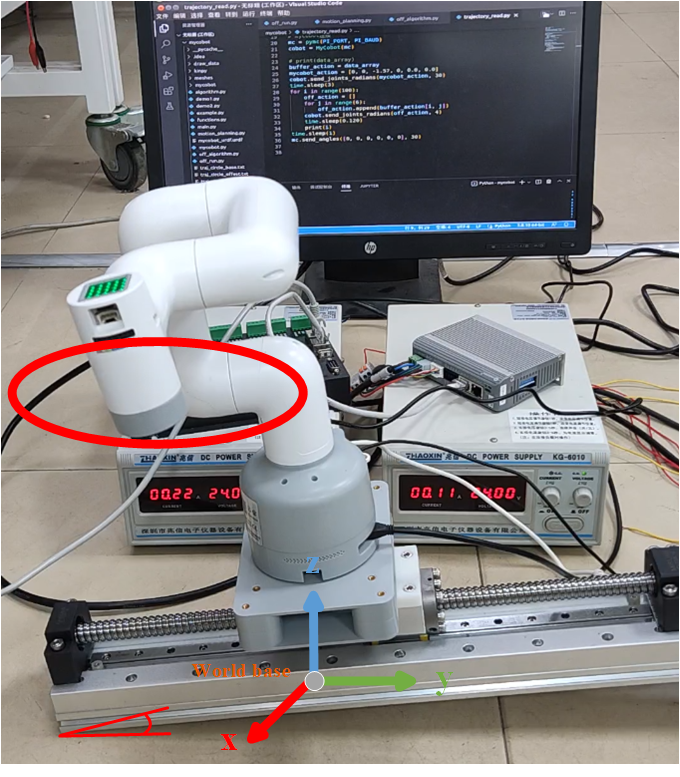}\\
			\centering{time=10s}
		\end{minipage}
	}
	\centering{(b)}
	\caption{Experimental results of end-effector trajectory tracking of a six-degree-of-freedom manipulator. (a) The base is not tilted. (b) The base is tilted.}\label{FIG_7}
\end{figure} 

As shown in Fig.~\ref{FIG_5}, the position and orientation errors of the end-effector under the proposed POMPTC scheme and PD control are compared. It can be seen that the proposed method has higher control accuracy than PD control. Subsequently, the tracking errors of the proposed NFTSM, PID, and ESO (Extended State Observer) \cite{article_31} methods in three coordinate axes of the Cartesian coordinate system are compared. As shown in Fig.~\ref{FIG_6}, the proposed method has faster convergence speed and smaller steady-state error than PID and ESO. Therefore, in terms of control accuracy, the proposed method is superior to PID and ESO methods.

From the analysis of the entire simulation process, it can be seen that the proposed method exhibits good control performance under changes in the base pose of the manipulator. It can stabilize the position error of the manipulator end-effector trajectory tracking within 1mm, and the orientation error within 0.01rad.

\subsection{Experiments}

\begin{figure}[!t]
	\centering
	\subfigure
	{
		\begin{minipage}[t]{.5\linewidth}
			\includegraphics[scale=0.21]{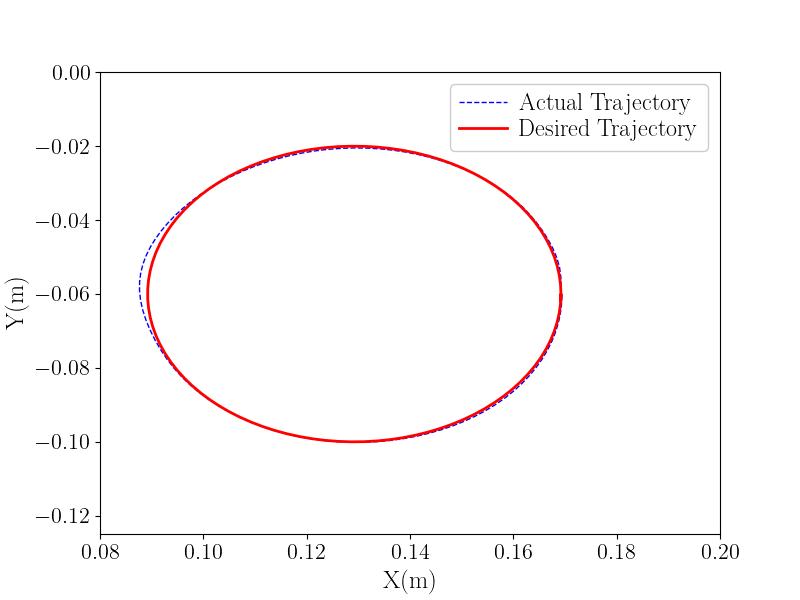}\\
			\centering{(a)}
		\end{minipage}
		\begin{minipage}[t]{.5\linewidth}
			\includegraphics[scale=0.21]{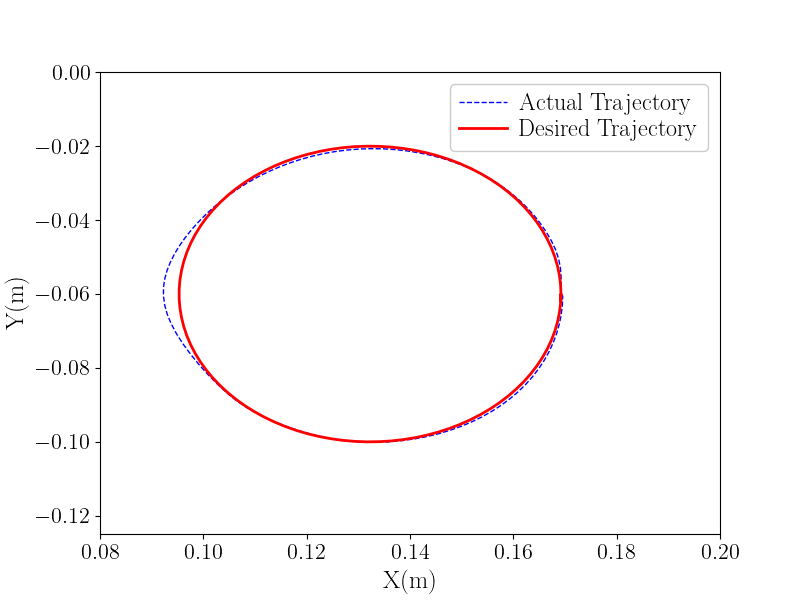}\\
			\centering{(b)}
		\end{minipage}		
	}
	\caption{Trajectory of the end-effector. (a) The base is not tilted. (b) The base is tilted.}\label{FIG_8}
\end{figure}

To further verify the feasibility and advantages of the proposed method, a guide rail and a myCobot 280 six-degree-of-freedom manipulator are used in the experiment. The specific experiments are shown in Fig.~\ref{FIG_7}. Under the conditions of horizontal placement of the base and tilting of the base, the base is reciprocated on the guide rail, while the end of the manipulator needs to track a circular trajectory in the world coordinate system. The manipulator base is fixed on the moving slider of the guide rail and moves with the slider, so tilting the guide rail can make the base tilt. In the case of tilting the base, with a tilt angle of about 0.21 rad, the end of the manipulator needs to consider the tilt angle and track the same trajectory as in the case of no tilt with the base. As shown in Fig.~\ref{FIG_8}, the trajectories of the manipulator under two conditions are displayed: without tilting the base and with tilting the base.

\begin{figure}[!t]
	\centering
	\subfigure
	{
		\begin{minipage}[t]{.33\linewidth}
			\includegraphics[scale=0.2]{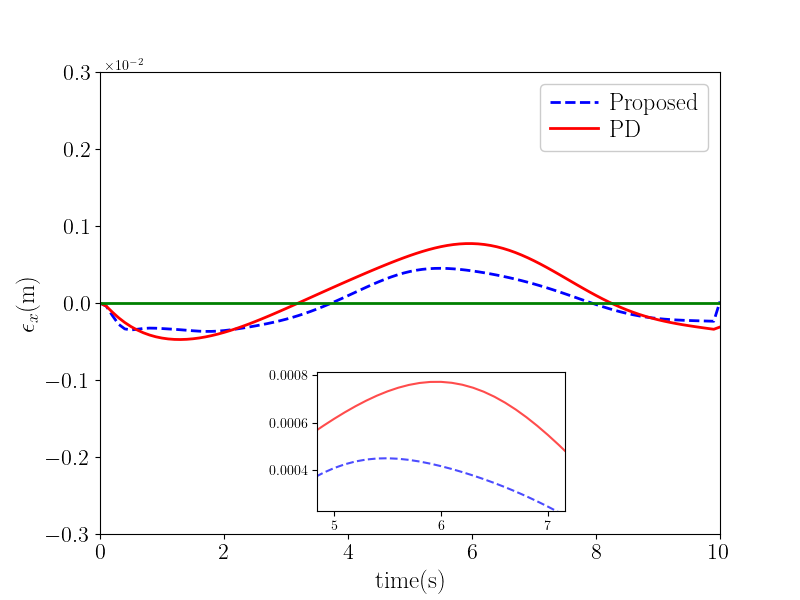}
		\end{minipage}
		\begin{minipage}[t]{.33\linewidth}
			\includegraphics[scale=0.22]{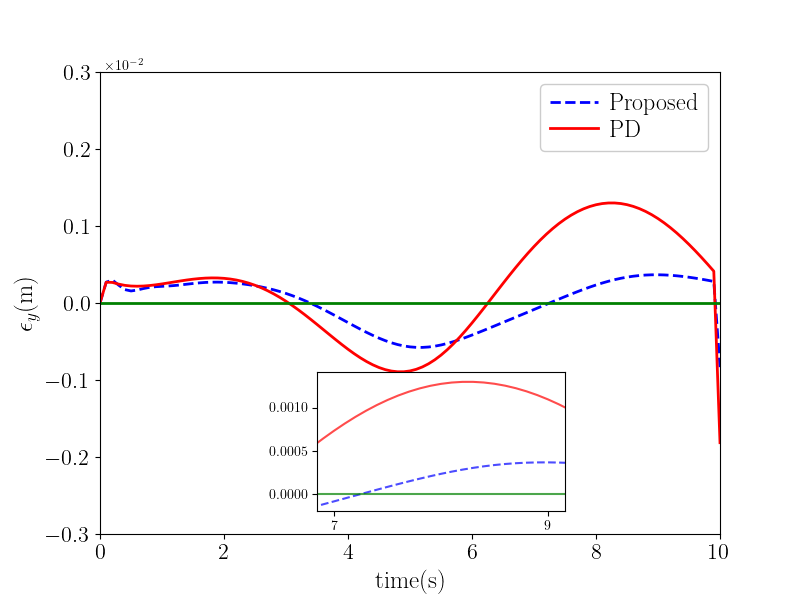}
		\end{minipage}		
		\begin{minipage}[t]{.33\linewidth}
			\includegraphics[scale=0.22]{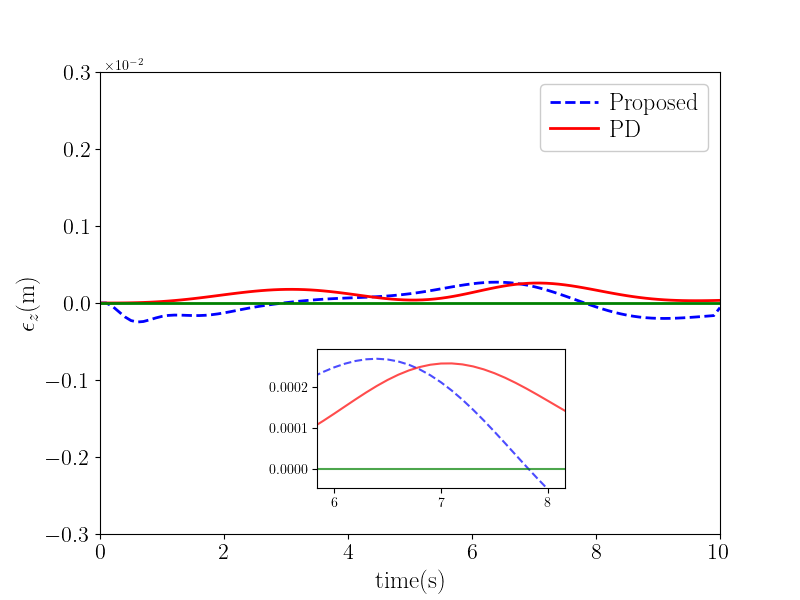}
		\end{minipage}
	}
	\centering{(a)}
	\subfigure
	{
		\begin{minipage}[t]{.33\linewidth}
			\includegraphics[scale=0.2]{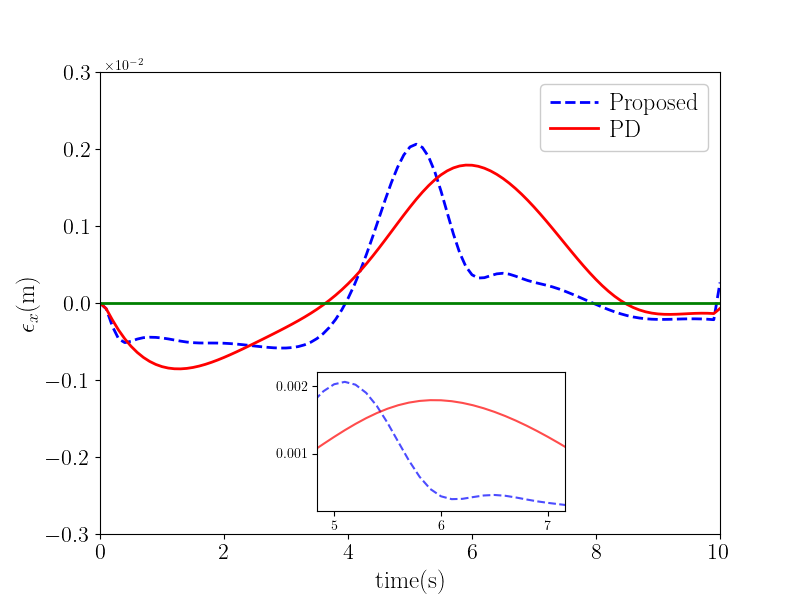}
		\end{minipage}
		\begin{minipage}[t]{.33\linewidth}
			\includegraphics[scale=0.2]{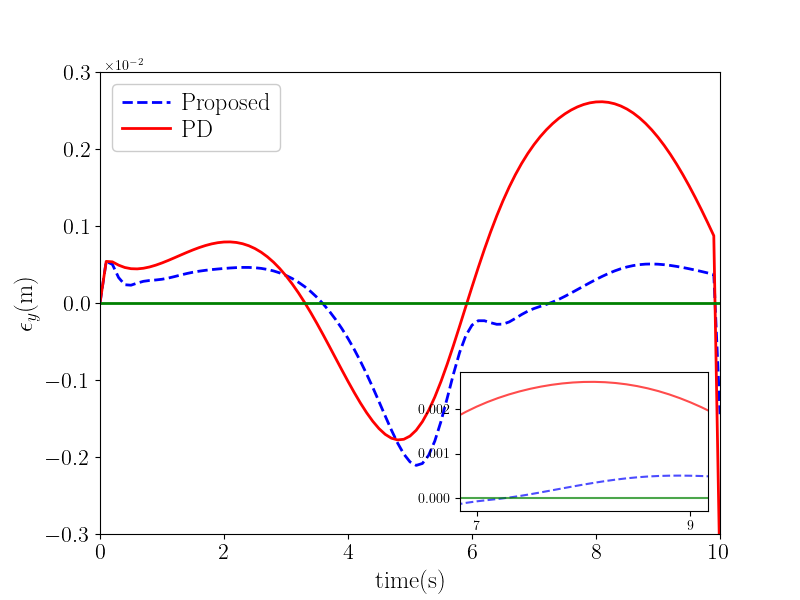}
		\end{minipage}		
		\begin{minipage}[t]{.33\linewidth}
			\includegraphics[scale=0.2]{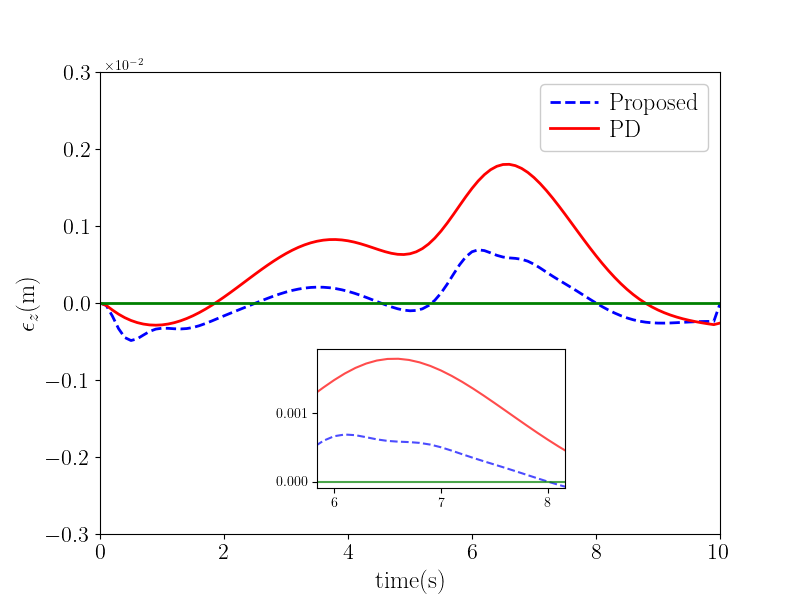}
		\end{minipage}
	}
	\centering{(b)}
	\caption{Experimental results of the end position error of the manipulator under different control algorithms. (a)The base is not tilted. (b)The base is tilted.}\label{FIG_9}
\end{figure}

Subsequently, the proposed method is compared with the PD control method. The specific content includes comparing the tracking errors of the proposed method and the PD method in the three coordinate axes of the Cartesian coordinate system. As shown in Fig.~\ref{FIG_9}, we find that under the disturbance caused by the motion of the base, regardless of whether the base is tilted or not, the proposed method has a faster convergence rate and smaller steady-state error. In addition, compared with the cases where the base is not tilted and tilted, the proposed method has a smaller error variation. Therefore, in terms of control accuracy and anti-disturbance ability, the proposed method is superior to PD control.

\section{CONCLUSION} \label{section:6}

For the MMs system, a new trajectory tracking control method combining NFTSM and MPC is proposed. This method can not only minimize the norm of three levels, but also achieve synchronous control of the position and orientation of the manipulator within a finite time. In addition, the proposed method can still achieve good control performance under the disturbance of base motion on the manipulator. In general, compared with other control methods, this method has the advantages of fast convergence, high control accuracy, and strong robustness. In the future, we will study the collaborative operation of redundant dual-arm robots with mobile bases.

\footnotesize{}
\bibliographystyle{Bibliography/IEEEtranTIE}
\bibliography{Bibliography/IEEEref}

\begin{thebibliography}{10}
\providecommand{\url}[1]{#1}
\csname url@samestyle\endcsname
\providecommand{\newblock}{\relax}
\providecommand{\bibinfo}[2]{#2}
\providecommand{\BIBentrySTDinterwordspacing}{\spaceskip=0pt\relax}
\providecommand{\BIBentryALTinterwordstretchfactor}{4}
\providecommand{\BIBentryALTinterwordspacing}{\spaceskip=\fontdimen2\font plus
\BIBentryALTinterwordstretchfactor\fontdimen3\font minus
  \fontdimen4\font\relax}
\providecommand{\BIBforeignlanguage}[2]{{%
\expandafter\ifx\csname l@#1\endcsname\relax
\typeout{** WARNING: IEEEtran.bst: No hyphenation pattern has been}%
\typeout{** loaded for the language `#1'. Using the pattern for}%
\typeout{** the default language instead.}%
\else
\language=\csname l@#1\endcsname
\fi
#2}}
\providecommand{\BIBdecl}{\relax}
\BIBdecl

\bibitem{article_1}
W.~Ye, Z.~Li, C.~Yang, J.~Sun, C.-Y. Su, and R.~Lu, ``Vision-based human
  tracking control of a wheeled inverted pendulum robot,'' \emph{IEEE
  Transactions on Cybernetics}, vol.~46,
  \href{http://dx.doi.org/10.1109/TCYB.2015.2478154}{DOI
  10.1109/TCYB.2015.2478154}, no.~11, pp. 2423--2434, 2016.

\bibitem{article_2}
D.~Zhou, M.~Shi, F.~Chao, C.-M. Lin, L.~Yang, C.~Shang, and C.~Zhou, ``Use of
  human gestures for controlling a mobile robot via adaptive cmac network and
  fuzzy logic controller,'' \emph{Neurocomputing}, vol. 282,
  \href{http://dx.doi.org/10.1016/j.neucom.2017.12.016}{DOI
  10.1016/j.neucom.2017.12.016}, pp. 218--231, MAR 22 2018.

\bibitem{article_28}
Y.~Leng, X.~Lin, G.~Huang, M.~Hao, J.~Wu, Y.~Xiang, K.~Zhang, and C.~Fu,
  ``Wheel-legged robotic limb to assist human with load carriage: An
  application for environmental disinfection during covid-19,'' \emph{IEEE
  Robotics and Automation Letters}, vol.~6,
  \href{http://dx.doi.org/10.1109/LRA.2021.3065197}{DOI
  10.1109/LRA.2021.3065197}, no.~2, pp. 3695--3702, 2021.

\bibitem{article_29}
Y.~Xie, X.~Zhang, S.~Zheng, C.~K. Ahn, and S.~Wang, ``Asynchronous h continuous
  stabilization of mode-dependent switched mobile robot,'' \emph{IEEE
  Transactions on Systems, Man, and Cybernetics: Systems}, vol.~52,
  \href{http://dx.doi.org/10.1109/TSMC.2021.3119054}{DOI
  10.1109/TSMC.2021.3119054}, no.~11, pp. 6906--6920, 2022.

\bibitem{article_32}
S.~Liu, H.~Chai, R.~Song, Y.~Li, Y.~Li, Q.~Zhang, P.~Fu, J.~Liu, and Z.~Yang,
  ``Contact force/motion hybrid control for a hydraulic legged mobile
  manipulator via a force-controlled floating base,'' \emph{IEEE/ASME
  Transactions on Mechatronics}, vol.~29,
  \href{http://dx.doi.org/10.1109/TMECH.2023.3323541}{DOI
  10.1109/TMECH.2023.3323541}, no.~3, pp. 2316--2326, 2024.

\bibitem{article_3}
G.~Peng, C.~Yang, W.~He, and C.~L.~P. Chen, ``Force sensorless admittance
  control with neural learning for robots with actuator saturation,''
  \emph{IEEE Transactions on Industrial Electronics}, vol.~67,
  \href{http://dx.doi.org/10.1109/TIE.2019.2912781}{DOI
  10.1109/TIE.2019.2912781}, no.~4, pp. 3138--3148, 2020.

\bibitem{article_4}
A.~Dietrich and C.~Ott, ``Hierarchical impedance-based tracking control of
  kinematically redundant robots,'' \emph{IEEE Transactions on Robotics},
  vol.~36, \href{http://dx.doi.org/10.1109/TRO.2019.2945876}{DOI
  10.1109/TRO.2019.2945876}, no.~1, pp. 204--221, 2020.

\bibitem{article_5}
J.~Yan, L.~Jin, Z.~Yuan, and Z.~Liu, ``Rnn for receding horizon control of
  redundant robot manipulators,'' \emph{IEEE Transactions on Industrial
  Electronics}, vol.~69, \href{http://dx.doi.org/10.1109/TIE.2021.3062257}{DOI
  10.1109/TIE.2021.3062257}, no.~2, pp. 1608--1619, 2022.

\bibitem{article_6}
G.~Peng, C.~L.~P. Chen, W.~He, and C.~Yang, ``Neural-learning-based force
  sensorless admittance control for robots with input deadzone,'' \emph{IEEE
  Transactions on Industrial Electronics}, vol.~68,
  \href{http://dx.doi.org/10.1109/TIE.2020.2991929}{DOI
  10.1109/TIE.2020.2991929}, no.~6, pp. 5184--5196, 2021.

\bibitem{article_7}
L.~Jin and Y.~Zhang, ``G2-type srmpc scheme for synchronous manipulation of two
  redundant robot arms,'' \emph{IEEE Transactions on Cybernetics}, vol.~45,
  \href{http://dx.doi.org/10.1109/TCYB.2014.2321390}{DOI
  10.1109/TCYB.2014.2321390}, no.~2, pp. 153--164, 2015.

\bibitem{article_8}
Z.~Xie, L.~Jin, X.~Du, X.~Xiao, H.~Li, and S.~Li, ``On generalized rmp scheme
  for redundant robot manipulators aided with dynamic neural networks and
  nonconvex bound constraints,'' \emph{IEEE Transactions on Industrial
  Informatics}, vol.~15, \href{http://dx.doi.org/10.1109/TII.2019.2899909}{DOI
  10.1109/TII.2019.2899909}, no.~9, pp. 5172--5181, 2019.

\bibitem{article_9}
Z.~Zhang, S.~Chen, X.~Zhu, and Z.~Yan, ``Two hybrid end-effector
  posture-maintaining and obstacle-limits avoidance schemes for redundant robot
  manipulators,'' \emph{IEEE Transactions on Industrial Informatics}, vol.~16,
  \href{http://dx.doi.org/10.1109/TII.2019.2922694}{DOI
  10.1109/TII.2019.2922694}, no.~2, pp. 754--763, 2020.

\bibitem{article_10}
J.~Fan, L.~Jin, Z.~Xie, S.~Li, and Y.~Zheng, ``Data-driven motion-force control
  scheme for redundant manipulators: A kinematic perspective,'' \emph{IEEE
  Transactions on Industrial Informatics}, vol.~18,
  \href{http://dx.doi.org/10.1109/TII.2021.3125449}{DOI
  10.1109/TII.2021.3125449}, no.~8, pp. 5338--5347, 2022.

\bibitem{article_11}
L.~Xiao and Y.~Zhang, ``Dynamic design, numerical solution and effective
  verification of acceleration-level obstacle-avoidance scheme for robot
  manipulators,'' \emph{Internationnal Journal of Systemes Science}, vol.~47,
  \href{http://dx.doi.org/10.1080/00207721.2014.909971}{DOI
  10.1080/00207721.2014.909971}, no.~4, pp. 932--945, MAR 11 2016.

\bibitem{article_12}
J.~Fan, L.~Jin, Z.~Xie, S.~Li, and Y.~Zheng, ``Data-driven motion-force control
  scheme for redundant manipulators: A kinematic perspective,'' \emph{IEEE
  Transactions on Industrial Informatics}, vol.~18,
  \href{http://dx.doi.org/10.1109/TII.2021.3125449}{DOI
  10.1109/TII.2021.3125449}, no.~8, pp. 5338--5347, 2022.

\bibitem{article_13}
V.~N. Katsikis, P.~S. Stanimirovi, S.~D. Mourtas, L.~Xiao, D.~Karabaevi, and
  D.~Stanujki, ``Zeroing neural network with fuzzy parameter for computing
  pseudoinverse of arbitrary matrix,'' \emph{IEEE Transactions on Fuzzy
  Systems}, vol.~30, \href{http://dx.doi.org/10.1109/TFUZZ.2021.3115969}{DOI
  10.1109/TFUZZ.2021.3115969}, no.~9, pp. 3426--3435, 2022.

\bibitem{article_14}
Y.~Shi, W.~Zhao, S.~Li, B.~Li, and X.~Sun, ``Novel discrete-time recurrent
  neural network for robot manipulator: A direct discretization technical
  route,'' \emph{IEEE Transactions on Neural Networks and Learning Systems},
  vol.~34, \href{http://dx.doi.org/10.1109/TNNLS.2021.3108050}{DOI
  10.1109/TNNLS.2021.3108050}, no.~6, pp. 2781--2790, 2023.

\bibitem{article_27}
Z.~Xu, S.~Li, X.~Zhou, S.~Zhou, T.~Cheng, and Y.~Guan, ``Dynamic neural
  networks for motion-force control of redundant manipulators: An optimization
  perspective,'' \emph{IEEE Transactions on Industrial Electronics}, vol.~68,
  \href{http://dx.doi.org/10.1109/TIE.2020.2970635}{DOI
  10.1109/TIE.2020.2970635}, no.~2, pp. 1525--1536, 2021.

\bibitem{article_15}
W.~Xu, A.~K. Junejo, Y.~Liu, and M.~R. Islam, ``Improved continuous fast
  terminal sliding mode control with extended state observer for speed
  regulation of pmsm drive system,'' \emph{IEEE Transactions on Vehicular
  Technology}, vol.~68, \href{http://dx.doi.org/10.1109/TVT.2019.2926316}{DOI
  10.1109/TVT.2019.2926316}, no.~11, pp. 10\,465--10\,476, 2019.

\bibitem{article_16}
F.-J. Lin, J.-C. Hwang, P.-H. Chou, and Y.-C. Hung, ``Fpga-based
  intelligent-complementary sliding-mode control for pmlsm servo-drive
  system,'' \emph{IEEE Transactions on Power Electronics}, vol.~25,
  \href{http://dx.doi.org/10.1109/TPEL.2010.2050907}{DOI
  10.1109/TPEL.2010.2050907}, no.~10, pp. 2573--2587, 2010.

\bibitem{article_17}
W.-T. Su and C.-M. Liaw, ``Adaptive positioning control for a lpmsm drive based
  on adapted inverse model and robust disturbance observer,'' \emph{IEEE
  Transactions on Power Electronics}, vol.~21,
  \href{http://dx.doi.org/10.1109/TPEL.2005.869729}{DOI
  10.1109/TPEL.2005.869729}, no.~2, pp. 505--517, 2006.

\bibitem{article_18}
F.~J. Lin, P.~H. Chou, Y.~C. Hung, and W.~M. Wang, ``Field-programmable gate
  array-based functional link radial basis function network control for
  permanent magnet linear synchronous motor servo drive system,'' \emph{Iet
  Electric Power Applications}, vol.~4,
  \href{http://dx.doi.org/10.1049/iet-epa.2009.0104}{DOI
  10.1049/iet-epa.2009.0104}, no.~5, pp. 357--372, May. 2010.

\bibitem{article_30}
L.~Jiang, S.~Wang, Y.~Xie, J.~Meng, S.~Zheng, X.~Zhang, and H.~Wu,
  ``Anti-disturbance direct yaw moment control of a four-wheeled autonomous
  mobile robot,'' \emph{IEEE Access}, vol.~8,
  \href{http://dx.doi.org/10.1109/ACCESS.2020.3025575}{DOI
  10.1109/ACCESS.2020.3025575}, pp. 174\,654--174\,666, 2020.

\bibitem{article_33}
Z.~Liu, Y.~Zhao, O.~Zhang, W.~Chen, J.~Wang, Y.~Gao, and J.~Liu, ``A novel
  faster fixed-time adaptive control for robotic systems with input
  saturation,'' \emph{IEEE Transactions on Industrial Electronics}, vol.~71,
  \href{http://dx.doi.org/10.1109/TIE.2023.3281701}{DOI
  10.1109/TIE.2023.3281701}, no.~5, pp. 5215--5223, 2024.

\bibitem{article_19}
H.~Wang, L.~Shi, Z.~Man, J.~Zheng, S.~Li, M.~Yu, C.~Jiang, H.~Kong, and Z.~Cao,
  ``Continuous fast nonsingular terminal sliding mode control of automotive
  electronic throttle systems using finite-time exact observer,'' \emph{IEEE
  Transactions on Industrial Electronics}, vol.~65,
  \href{http://dx.doi.org/10.1109/TIE.2018.2795591}{DOI
  10.1109/TIE.2018.2795591}, no.~9, pp. 7160--7172, 2018.

\bibitem{article_20}
Y.~Wu, X.~Yu, and Z.~Man, ``Terminal sliding mode control design for uncertain
  dynamic systems,'' \emph{System and Control Letters}, vol.~34,
  \href{http://dx.doi.org/10.1016/S0167-6911(98)00036-X}{DOI
  10.1016/S0167-6911(98)00036-X}, no.~5, pp. 281--287, JUL 27 1998.

\bibitem{article_21}
S.~Yu, X.~Yu, B.~Shirinzadeh, and Z.~Man, ``Continuous finite-time control for
  robotic manipulators with terminal sliding mode,'' \emph{Automatica},
  vol.~41, \href{http://dx.doi.org/10.1016/j.automatica.2005.07.001}{DOI
  10.1016/j.automatica.2005.07.001}, no.~11, pp. 1957--1964, Nov. 2005.

\bibitem{article_22}
L.~Yang and J.~Yang, ``Nonsingular fast terminal sliding-mode control for
  nonlinear dynamical systems,'' \emph{Internationnal Journal of Robust and
  Nonlinear Control}, vol.~21, \href{http://dx.doi.org/10.1002/rnc.1666}{DOI
  10.1002/rnc.1666}, no.~16, pp. 1865--1879, NOV 10 2011.

\bibitem{article_23}
J.~Zheng, H.~Wang, Z.~Man, J.~Jin, and M.~Fu, ``Robust motion control of a
  linear motor positioner using fast nonsingular terminal sliding mode,''
  \emph{IEEE/ASME Transactions on Mechatronics}, vol.~20,
  \href{http://dx.doi.org/10.1109/TMECH.2014.2352647}{DOI
  10.1109/TMECH.2014.2352647}, no.~4, pp. 1743--1752, 2015.

\bibitem{article_24}
M.~Van, M.~Mavrovouniotis, and S.~S. Ge, ``An adaptive backstepping nonsingular
  fast terminal sliding mode control for robust fault tolerant control of robot
  manipulators,'' \emph{IEEE Transactions on Systems, Man, and Cybernetics:
  Systems}, vol.~49, \href{http://dx.doi.org/10.1109/TSMC.2017.2782246}{DOI
  10.1109/TSMC.2017.2782246}, no.~7, pp. 1448--1458, 2019.

\bibitem{article_25}
J.~Ni, L.~Liu, C.~Liu, X.~Hu, and S.~Li, ``Fast fixed-time nonsingular terminal
  sliding mode control and its application to chaos suppression in power
  system,'' \emph{IEEE Transactions on Circuits and Systems II: Express
  Briefs}, vol.~64, \href{http://dx.doi.org/10.1109/TCSII.2016.2551539}{DOI
  10.1109/TCSII.2016.2551539}, no.~2, pp. 151--155, 2017.

\bibitem{article_26}
H.~M. La, R.~Lim, and W.~Sheng, ``Multirobot cooperative learning for predator
  avoidance,'' \emph{IEEE Transactions on Control Systems Technology}, vol.~23,
  \href{http://dx.doi.org/10.1109/TCST.2014.2312392}{DOI
  10.1109/TCST.2014.2312392}, no.~1, pp. 52--63, 2015.

\bibitem{article_31}
C.~Ren, X.~Li, X.~Yang, and S.~Ma, ``Extended state observer-based sliding mode
  control of an omnidirectional mobile robot with friction compensation,''
  \emph{IEEE Transactions on Industrial Electronics}, vol.~66,
  \href{http://dx.doi.org/10.1109/TIE.2019.2892678}{DOI
  10.1109/TIE.2019.2892678}, no.~12, pp. 9480--9489, 2019.

\end{thebibliography}

\normalsize{}
\clearpage
\appendices

\section{FINITE-TIME STABILITY ANALYSIS}\label{app:finite_time}
\begin{thm}
	For the POMPTC scheme \eqref{eq10}, starting from any initial value $v(0)=[z^{\rm T}(0),\varphi^{\rm T}(0)]^{\rm T}$, the variable $v(t)=[z^{\rm T}(t),\varphi^{\rm T}(t)]^{\rm T}$ of \eqref{eq19} globally converge to the theoretical solution $v^*(t)=[z^{*{\rm T}}(t),\varphi^{*{\rm T}}(t)]^T$, where $z^{*}(t)$ represents the joint velocity increment of the manipulator during the sampling period $t$.
\end{thm}

\begin{IEEEproof}
	The Lyapunov function is defined as
	
	\begin{equation}
		\label{eq21}
		\mathcal{F}(t)=h^{\rm T}(t)h(t), 
	\end{equation}
	where $\mathcal{F}(t)$ is positive semi-definite. Then, the derivative of \eqref{eq21} with respect to $t$ is expressed as
	
	\begin{equation}
		\dot{\mathcal{F}}(t)=h^{\rm T}(t)\dot{h}(t)=-\mu{h^{\rm T}(t)}\Omega(h(t)), 
	\end{equation}
	since $\Omega(\cdot)$ is an odd function and monotonically increasing, and $\mu > 0$, we have $\dot{\mathcal{F}}(t) \leq 0$. According to Lyapunov stability theory, it can be inferred that $h(t)$ converges to zero. Therefore, starting from any initial value $v(0)=[z^{\rm T}(0),\varphi^{\rm T}(0)]^{\rm T},$ the variable $v(t)=[z^{\rm T}(t),\varphi^{\rm T}(t)]^{\rm T}$ of \eqref{eq20} globally converges to the theoretical solution $v^*(t)=[z^{*{\rm T}}(t),\varphi^{*{\rm T}}(t)]^{\rm T}$, where $z^{*}(t)$ represents the joint velocity increment of the manipulator during the sampling period $t$.
	
	The proof is completed.
\end{IEEEproof}

\begin{thm}
	For the POMPTC scheme \eqref{eq10}, the joint velocity increment $z^* (t)$ of the manipulator during the sampling period $t$ is derived within a finite time $t_f$ to achieve finite-time control, where $t_f$ is expressed as
\end{thm}

\begin{equation*}
	t_f\le{\frac{2|h_{max}(0)|^{1-\kappa}}{\mu(1-\kappa)}},
\end{equation*}
among them, $h_{max}(0)$ comes from the initial value of the error function $h(0)$ , which represents the element with the largest absolute value.

\begin{IEEEproof}
	Define $h_{max}(t)$ as the element of $h(t)$ with $h_{max}(0)$. From $\dot{h}(t)=-\mu\Omega(h(t))$, the following equation can be obtained:
	
	\begin{equation*}
		\dot{h}_i(t)=-\mu\omega(h_i(t)),
	\end{equation*}
	where $\dot{h}_i(t)$ represents the time derivative of $h_i(t)$. Based on the sign of $h_{max}(0)$, the analysis can be divided into the following three cases.
	
	When $h_{max}(0)>0$, for all appropriate elements $i$, $h_{max}(0)\ge{h_i(0)}$. According to the comparison theorem, for all appropriate elements $i$, $h_{max}(t)\ge{h_i(t)}$ when $t\ge0$. Similarly, for all appropriate elements $i$, when $t\ge0$, $-h_{max}(t)\le{h_i(t)}$. Therefore, as time $t$ increases, for all appropriate elements $i$, $-h_{max}(t)\le{h_i(t)}\le{h_{max}(t)}$, which indicates that $h_i(t)$ converges to zero as $h_{max}(t)$ tends to zero. That is to say, the convergence time of the FTCND model \eqref{eq20} is bounded by $t_{fmax}$, where $t_{fmax}$ is the dynamic convergence time of $h_{max}(t)$, that is, $t_f\le{t_{fmax}}$. As for $t_{fmax}$,
	
	\begin{equation*}
		\dot{h}_{max}(t)=-\mu{\omega(h_{max}(t))}.
	\end{equation*}
	
	Then, a Lyapunov candidate function is defined as $Y(t)=|h_{max}(t)|^2$, and its time derivation is as follows.
	
	\begin{equation*}
		\begin{aligned}
			\dot{Y}(t)&=-2\mu{h_{max}}(t)\omega(h_{max}(t)) \\
			&=-\mu(\lambda|h_{max}(t)|^{\kappa+1}+\lambda|h_{max}(t)|^{(1/\kappa)+1})+\zeta{h_{max}^2}(t)) \\
			&\le-\mu\lambda|h_{max}(t)|^{\kappa+1})=-\mu\lambda{Y^{(\kappa+1)/2}}(t)
		\end{aligned}
	\end{equation*}
	
	The proof is completed.
\end{IEEEproof}

\section{STABILITY ANALYSIS} \label{app:stability}
\begin{IEEEproof}
	The derivative of the sliding mode surface is expressed as
	
	\begin{equation}
		\label{eq30}
		\dot{s}=\widetilde{q}_2+\alpha{r_1}|\widetilde{q}_1|^{r_1-1}\widetilde{q}_2+\beta{r_2}|\widetilde{q}_2|^{r_2-1}\dot{\widetilde{q}}_2,
	\end{equation}
	
	Substituting \eqref{eq24} into \eqref{eq30} , $\dot{s}$ can be expressed as 
	
	\begin{equation}
		\label{eq31}
		\begin{aligned}
			\dot{s}=&\widetilde{q}_2+\alpha{r_1}|\widetilde{q}_1|^{r_1-1}\widetilde{q}_2+\beta{r_2}|\widetilde{q}_2|^{r_2-1} \\
			&(F(\widetilde{q})+M^{-1}(q)\tau+D(\widetilde{q},\dot{\widetilde{q}})+B(\widetilde{q},\dot{\widetilde{q}})), 
		\end{aligned}
	\end{equation}
	
	By substituting the control inputs from \eqref{eq28} and \eqref{eq29} into \eqref{eq31} , $\dot{s}$ can expressed as
	\begin{equation}
		\label{eq32}
		\begin{aligned}
			\dot{s}=&\widetilde{q}_2+\alpha{r1}|\widetilde{q}_1|^{r_1-1}\widetilde{q}_2+\beta{r_2}|\widetilde{q}_2|^{r_2-1} \\
			&\cdot(-u_{eq}-u_{sw}+F(\widetilde{q})+D(\widetilde{q},\dot{\widetilde{q}})+B(\widetilde{q},\dot{\widetilde{q}})) \\
					=&\widetilde{q}_2+\alpha{r1}|\widetilde{q}_1|^{r_1-1}\widetilde{q}_2+\beta{r_2}|\widetilde{q}_2|^{r_2-1}(-u_{sw}+D(\widetilde{q},\dot{\widetilde{q}})+B(\widetilde{q},\dot{\widetilde{q}})) \\
					&+\beta{r_2}|\widetilde{q}_2|^{r_2-1}\left[-\frac{|\widetilde{q}_2|^{2-r_2}sat(\widetilde{q}_2)}{\beta{r_2}}(1+\alpha{r_1}|\widetilde{q}_1|^{r_1-1})\right] \\
					=&\widetilde{q}_2+\alpha{r1}|\widetilde{q}_1|^{r_1-1}\widetilde{q}_2+\beta{r_2}|\widetilde{q}_2|^{r_2-1}(-u_{sw}+D(\widetilde{q},\dot{\widetilde{q}})+B(\widetilde{q},\dot{\widetilde{q}})) \\
					&-|\widetilde{q}_2|sat(\widetilde{q}_2)(1+\alpha{r_1}|\widetilde{q}_1|^{r_1-1}) \\
			=&\beta{r_2}|\widetilde{q}_2|^{r_2-1}(-c_1|s|^{r_3}sat(s)-c_2s+D(\widetilde{q},\dot{\widetilde{q}})+B(\widetilde{q},\dot{\widetilde{q}})).
		\end{aligned}
	\end{equation}
	
	Let $d=\beta{r_2}|\widetilde{q}_2|^{r_2-1}$, and $d>0$. The Lyapunov function is defined as
	
	\begin{equation}
		\label{eq33}
		V=\frac{1}{2}ss^{{\rm T}}.
	\end{equation}
	
	The derivative of $V$ can be written as
	
	\begin{equation}
		\label{eq34}
		\begin{aligned}
			\dot{V}=&s\dot{s}=-c_1d|s|^{r_3}s\cdot{sat(s)}-c_2ds^2+d(D(\widetilde{q},\dot{\widetilde{q}})+B(\widetilde{q},\dot{\widetilde{q}}))s \\
			\le&-c_1d|s|^{r_3}|s|-c_2ds^2+d(D(\widetilde{q},\dot{\widetilde{q}})+B(\widetilde{q},\dot{\widetilde{q}}))|s| \\
			=&-2^{\frac{r_3+1}{2}}d \left(c_1-\frac{D(\widetilde{q},\dot{\widetilde{q}})+B(\widetilde{q},\dot{\widetilde{q}})}{|s|^{r_3}} \right)V^{\frac{r_3+1}{2}}-2c_2dV.
		\end{aligned}
	\end{equation}
	
	Therefore, according to Assumption~\ref{assump:1}, when $c_1>\frac{D(\widetilde{q}.\dot{\widetilde{q}})+B(\widetilde{q},\dot{\widetilde{q}})}{|s|^{r_3}}$ is satisfied, the position tracking error converges to zero in finite time. However, the lumped disturbances is uncertain. To ensure the stability and robustness of the system, $c_1$ needs to be large enough, otherwise it will cause system chattering.
	
	The proof is completed.
\end{IEEEproof}

\section{JOINT CONSTRAINTS}\label{app:constraints}

The joint constraints are given in Table~\ref{table_1}.

	\begin{table}[htbp]
		\renewcommand{\arraystretch}{1.2}
		\caption{Joint Limit}
		\centering
		\label{table_1}
		\centering
		\resizebox{0.55\linewidth}{!}{
			\begin{tabular}{c c}
				\hline\hline \\[-2mm]
				\multicolumn{1}{c}{Parameters} & \multicolumn{1}{c}{Value}\\[1.6ex] \hline
				$\eta_{B_{1U}}$ & $[1, 1, 3]~{\rm m}$ \\
				$\eta_{B_{2U}}$ & $[1, 1, 1]~{\rm rad}$ \\
				$\eta_{B_{1L}}$ & $[-1, -1, -1]~{\rm m}$ \\
				$\eta_{B_{2L}}$ &$[-1, -1, -1]~{\rm rad}$ \\
				$q_{mU}$ & $[2.5, 1.70, 2.5, -0.07, 2.5, 3.75, 2.5]^{\rm T}~{\rm rad}$ \\
				$q_{mL}$ & $[-2.5, -1.70, -2.5, -3.07, -2.5, -0.01, -2.5]^{\rm T}~\rm{rad}$ \\
				$\dot{\eta}_{B1U}$ & $[1, 1, 3]~{\rm m/s}$ \\
				$\dot{\eta}_{B2U}$ & $[1, 1, 1]~{\rm rad/s}$ \\
				$\dot{q}_{mU}$ & $[2, 2, 2, 2, 2.5, 2.5, 2.5]^{\rm T}~{\rm rad/s}$ \\
				$\dot{q}_{mL}$ & $[-2, -2, -2, -2, -2.5, -2.5, -2.5]^{\rm T}~{\rm rad/s}$ \\
				$\dot{\eta}_{B_{1L}}$ & $[-1, -1, -1]~{\rm m/s}$ \\
				$\dot{\eta}_{B_{2L}}$ & $[-1, -1, -1]~{\rm rad/s}$ \\
				$\ddot{\eta}_{B_{1U}}$ & $[1, 1, 1]~{\rm m/s^2}$ \\
				$\ddot{\eta}_{B_{2U}}$ & $[1, 1, 1]~{\rm rad/s^2}$ \\
				$\ddot{q}_{mU}$ & $[15, 7, 10, 12, 15, 20, 20]^{\rm T}~{\rm rad/s^2}$ \\
				$\ddot{q}_{mL}$ & $[-15, -7, -10, -12, -15, -20, -20]^{\rm T}~{\rm rad/s^2}$ \\
				$\ddot{\eta}_{B_{1L}}$ & $[-1,-1,-1]~{\rm m/s^2}$ \\
				$\ddot{\eta}_{B_{2L}}$ & $[-1, -1, -1]~{\rm rad/s^2}$ \\
				\hline\hline
			\end{tabular}
		}
	\end{table}

\end{document}